\documentclass[10pt,twocolumn,letterpaper]{article}
\usepackage[pagenumbers]{cvpr}
\usepackage[T1]{fontenc}

\usepackage{xurl}
\usepackage{tabularx}
\definecolor{linkblue}{rgb}{0.21,0.49,0.74}
\usepackage[pagebackref,breaklinks,colorlinks,allcolors=linkblue]{hyperref}
\usepackage{placeins}
\usepackage{amsthm}
\newcounter{proposition}
\newenvironment{proposition}[1][]{%
  \par\addvspace{\topsep}%
  \refstepcounter{proposition}%
  \noindent{\bfseries Proposition~\theproposition}%
  \if\relax\detokenize{#1}\relax\else\ (#1)\fi
  {\bfseries.}\hskip 5pt plus 1pt minus 1pt\ignorespaces}%
  {\par\addvspace{\topsep}}
\renewcommand{\eqref}[1]{\textup{(\ref{#1})}}
\definecolor{lightred}{RGB}{255,200,200}
\newcommand{\paperTitle}{SGMR: Learning Sparse Geometric Motion Representations for Robust Robotic Manipulation}
\newcommand{\paperAuthors}{Hu Fu, Junjie Huang, Fang Li, Zhixiang Duan, Teng Chen, Pengyi Li,
  Fangzhen Li, Wei Li, Bing Wang, Guang Chen, Yihua Tan, Zehan Zhang, Hangjun Ye}
\author{%
  {\fontsize{11}{13}\selectfont
  Hu Fu\textsuperscript{1,2,3*}\quad Junjie Huang\textsuperscript{3,4*}\quad
  Fang Li\textsuperscript{3,4}\quad Zhixiang Duan\textsuperscript{3,4}\quad
  Teng Chen\textsuperscript{3,4}\quad Pengyi Li}\\
  {\fontsize{11}{13}\selectfont
  Fangzhen Li\textsuperscript{3,4\textdagger}\enspace Wei Li\textsuperscript{3,4}\enspace
  Bing Wang\textsuperscript{3,4}\enspace Guang Chen\textsuperscript{3,4}\enspace
  Yihua Tan\textsuperscript{1,2\textdaggerdbl}\enspace
  Zehan Zhang\textsuperscript{3,4\textdagger\textdaggerdbl}\enspace Hangjun Ye\textsuperscript{3,4}}\\[4pt]
  {\small \textsuperscript{1}\,School of Artificial Intelligence and Automation, Huazhong University of Science and Technology, Wuhan}\\
  {\small \textsuperscript{2}\,National Key Laboratory of Multispectral Information Intelligent Processing Technology}\\
  {\small \textsuperscript{3}\,Xiaomi EV\qquad
          \textsuperscript{4}\,Beijing Key Laboratory of Intelligent Driving Foundation Models and Simulation}%
}
\hypersetup{pdfauthor={\paperAuthors}}

  \title{\paperTitle}
  \hypersetup{pdftitle={\paperTitle}}

\begin{document}

\maketitle
\begingroup
  \renewcommand{\thefootnote}{}
  \footnotetext[0]{\textsuperscript{*}\,Co-first authors.\quad
    \textsuperscript{\textdagger}\,Co-project leaders.\quad
    \textsuperscript{\textdaggerdbl}\,Corresponding authors.}
\endgroup
\begin{abstract}
Vision-language-action models inherit visual-semantic knowledge from pretrained vision-language models, yet remain vulnerable to distribution shifts in observations, instructions, and manipulation conditions. A key limitation is that pretrained VLMs lack explicit spatial awareness required for manipulation. Recent methods mitigate this limitation by aligning VLA representations with geometric teacher features. However, directly injecting such supervision into shared visual representations may severely interfere with the visual-semantic priors inherited from pretraining. Moreover, static spatial representations fail to capture how actions change subsequent observations and robot states, which restricts the adaptation of subsequent actions under perturbations.
To alleviate these issues, we introduce \textbf{SGMR}, a framework for learning \emph{\textbf{S}parse \textbf{G}eometric \textbf{M}otion \textbf{R}epresentations} for robust robotic manipulation that couples \textbf{Sparse Spatial Query Distillation} with an \textbf{Action Response Residual Flow}. To enhance spatial awareness while alleviate interference with pretrained visual-semantic priors, sparse spatial query distillation uses two independent query transformers to distill dense spatial knowledge from frozen VGGT and SAM3 teachers into compact tokens that give the action expert complementary world-grounded context. To support adaptive action generation, the action response residual flow relates recent executed actions to the subsequent changes in robot states and spatial tokens, which carry information about the current visual-motor relationship, and refine the next action chunk through a residual flow.
Experiments on LIBERO and VLABench, together with zero-shot evaluations on LIBERO-Plus and LIBERO-PRO, demonstrate that SGMR achieves outstanding manipulation performance while generalizing robustly across diverse perturbations. On a Language-extended Benchmark, SGMR achieves higher success rates under instruction rewrites and greater representational similarity to the pretrained VLM than both the baseline and direct feature alignment.

\end{abstract}

\section{Introduction}
\label{sec:intro}

\begin{figure}[t]
\centering
\includegraphics[width=\linewidth]{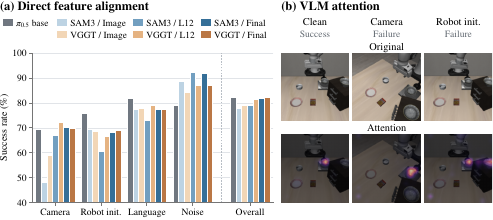}
\caption{\textbf{Perturbation Analysis.}
(a) LIBERO-Plus success rates with direct feature alignment.
(b) Initial VLM attention for successful clean and failed perturbed runs.}
\label{fig:intro-evidence}
\end{figure}

Vision-language-action (VLA) models leverage the rich visual-semantic priors of pretrained vision-language models (VLMs) for robotic control~\citep{kim2024openvla,black2024pi_0,intelligence2025pi_}. Despite excellent in-distribution performance, their robustness to shifts in visual observations, language instructions, and manipulation conditions remains limited, posing a major challenge to broader deployment~\citep{fei2025libero,zhou2025libero}. A key limitation is that VLM pretraining primarily emphasizes semantic understanding rather than the fine-grained geometric and spatial reasoning required for precise manipulation~\citep{li2026spatial}. Besides, policies driven primarily by current observations do not explicitly model how recent actions affect subsequent visual observations and proprioceptive states, potentially restricting their ability to adapt subsequent actions under perturbations.

Existing methods enhance VLA spatial awareness through geometric inputs~\citep{sun2025geovla}, spatial conditioning~\citep{qu2025spatialvla,jiang2025you}, and spatial representation learning~\citep{zhang2025spatial,feng2026spatial}. Among these approaches, distilling features from pretrained 3D foundation models, as a strategy for spatial representation learning, is particularly promising. It transfers rich geometric priors into intermediate policy representations without requiring explicit 3D observations or intermediate depth estimates, thereby reducing dependence on their availability and quality~\citep{li2026spatial}. However, directly aligning VLA visual features with geometric teacher representations may severely interfere with the visual-semantic structure inherited from VLM pretraining. Since the same representations also support semantic recognition and instruction grounding, improving geometric alignment alone does not necessarily yield more robust control.

To examine how direct feature alignment affects the pretrained VLM, we follow Spatial Forcing~\citep{li2026spatial} and fine-tune $\pi_{0.5}$ on LIBERO with an auxiliary loss that aligns its visual features at the image encoder, an intermediate VLM layer, or the final VLM layer with frozen VGGT or SAM3 features, and evaluate each model zero-shot on LIBERO-Plus. As seen in Figure~\ref{fig:intro-evidence}(a), alignment consistently improves robustness to visual noise but degrades language and robot-state perturbations. For example, aligning SAM3 features at layer 12 raises success under visual noise from 79.08\% to 92.50\% and lowers it under robot-state perturbations from 75.74\% to 60.71\%. This reveals a trade-off: \emph{imposing teacher supervision on shared VLA representations improves robustness to visual perturbations but may interfere with the pretrained visual-semantic structure that supports generalization along other dimensions}.

Moreover, static spatial representations do not capture the visual and proprioceptive consequences of recent actions. In Figure~\ref{fig:intro-evidence}(b), $\pi_{0.5}$ still attends to the target object under camera-viewpoint and robot-state perturbations but fails the task, which is consistent with a difficulty in translating target information into actions suited to the changed conditions. This motivates conditioning action generation on how the robot and the scene responded to recent actions.

To alleviate these challenges, we introduce \textbf{SGMR}, a framework for learning \emph{sparse geometric motion representations} for robust robotic manipulation. Specifically, to improve spatial awareness while alleviate interference with pretrained visual-semantic priors, \textbf{Sparse Spatial Query Distillation} utilizes two independent query transformers to compress dense geometric and semantic knowledge from frozen VGGT~\citep{wang2025vggt} and SAM3~\citep{carion2026sam} into compact tokens that provide spatial context to the action expert. Furthermore, to enable action adjustment under perturbations, the \textbf{Action Response Residual Flow} infers the current visual-motor relationship from recent actions and their subsequent effects on robot states and query representations, which adaptively refines the generation of the next action chunk through a residual flow ordinary differential equation (ODE).

Our main contributions are summarized as follows:
\begin{itemize}
\item We identify two challenges in robust VLA control: direct feature alignment will severely interfere with pretrained visual-semantic priors, while static spatial representations fail to model the visual-motor relationship under perturbations.
\item We propose \textbf{SGMR}, which couples \textbf{Sparse Spatial Query Distillation} with \textbf{Action Response Residual Flow}. The former compresses dense teacher knowledge into compact token states with independent query transformers, so the shared VLM tokens are not matched to teacher features, which alleviates interference with pretrained visual-semantic priors. The latter uses recent action responses to adaptively refine action generation.
\item We evaluate SGMR on LIBERO and VLABench and test it zero-shot on LIBERO-Plus and LIBERO-PRO. It achieves the best average results among the compared methods, showing that its spatial and temporal designs improve both manipulation performance and robustness to distribution shifts. The real-robot experiments further show its effectiveness in the real world.
\item To measure its effect on the pretrained VLM, we build a language-perturbation benchmark, on which SGMR is more robust to rewritten instructions than direct feature alignment and matches the baseline.
\end{itemize}

\section{Related Work}
\label{sec:related}

\paragraph{VLA policies and temporal context.}
VLAs adapt pretrained VLMs to control with discrete action tokens~\citep{brohan2023rt,kim2024openvla} or flow-matching action experts~\citep{black2024pi_0,intelligence2025pi_,zheng2026x}. Reasoning policies plan with goal images~\citep{zhao2025cot}, action references~\citep{zhong2026acot} or latent plans~\citep{huang2026fast,luo2026last}, while others restructure action generation with a latent action world model~\citep{bi2026motus}, geometry-aware latent actions for pretraining~\citep{liu2026gala}, world-flow and ego-trajectory representations~\citep{liu2026umr} or a geometric transport controller that hands over to local VLA \mbox{policies}~\citep{nguyen2026glovla}. Policies conditioned on history use multi-frame features~\citep{li2026towards}, visual traces~\citep{zheng2025tracevla}, memory banks~\citep{shi2026memoryvla}, keyframe-triggered reasoning memories~\citep{li2026trm} or recurrent belief states~\citep{xiao2026ava}. Closest to our work, \mbox{MotionVLA} encodes recent motion as trajectory-field tokens~\citep{yuan2026motionvla} and \mbox{OptimusVLA} encodes executed actions to keep successive chunks consistent~\citep{li2026global}. SGMR instead pairs executed actions with the subsequent changes in robot state and query features and uses these pairs to correct the velocity field of the next chunk.

\paragraph{Spatial representations.}
Geometry enters VLAs through 3D encoders~\citep{sun2025geovla,xia20263dvla}, 3D position encodings~\citep{qu2025spatialvla}, camera-aware observations or actions~\citep{jiang2025you,heo2026anycamvla,team2026one} and differentiable renderings that couple 3D and 2D VLAs~\citep{zhao2026localizing}, while active-perception policies select viewpoints or zoom in~\citep{liu2026activevla,liu2026sapave}. Other methods align VLA features with spatial teachers~\citep{li2026spatial}, fuse a separate spatial model~\citep{zhang2025spatial}, pretrain with 3D labels from human videos~\citep{feng2026spatial}, or predict future latent or 3D states~\citep{zhang2026dreamvla,sun2026vla,wang2026lamp,qian2026geopredict,xu2026action}. 3DThinkVLA aligns intermediate VLM features with a 3D foundation model during co-training~\citep{shi2026thinkvla}, StrataVLA lets selected layers query a frozen geometry model through adapters at inference~\citep{cui2026stratavla}. In SGMR, separate query transformers reconstruct the current teacher features and condition the action expert, so the shared VLM tokens are never matched to the teachers, which are dropped at inference.

\paragraph{Fine-tuning the pretrained VLM.}
Knowledge insulation blocks action-expert gradients into the VLM~\citep{driess2026knowledge}, MAPS schedules module-wise proximity to the pretrained weights~\citep{huang2026maps}, InstructVLA and Evo-1 co-train with instruction data or align the action module in stages to limit the loss of vision-language ability~\citep{yang2026vision,lin2026evo}, and MergeVLA limits interference between merged skills~\citep{fu2026mergevla}. SGMR keeps the VLM fully trainable and changes where spatial supervision enters.

\section{Preliminaries}
\label{sec:setup}

We consider language-conditioned robotic manipulation. At time step $t$, given multi-view RGB observations $I_t$, proprioceptive state $s_t$, and instruction $l$, the policy predicts an action chunk
\begin{equation}
A_t=(a_t,\ldots,a_{t+H-1})\in\mathbb{R}^{H\times d_a},
\end{equation}
where $H$ and $d_a$ denote the prediction horizon and action dimension, respectively. We adopt a flow-matching VLA policy that consists of a pretrained VLM backbone and an action expert. The VLM provides multi-modal context for the flow-matching action expert.

During training, we interpolate the demonstrated action $A_t$ with Gaussian noise $\epsilon\sim\mathcal{N}(0,\mathbf{I})$ as $z_\tau=(1-\tau)A_t+\tau\epsilon$, with $\tau\in[0,1]$, and optimize
\begin{equation}
\mathcal{L}_{\mathrm{flow}}(v_\theta) =
\mathbb{E}\left[
\frac{1}{Hd_a}
\left\|
v_\theta(z_\tau,\tau\mid I_t,s_t,l)-(\epsilon-A_t)
\right\|_F^2
\right].
\label{eq:flow}
\end{equation}
At inference, starting from $z_1=\epsilon$, the predicted action chunk $\widehat{A}_t=z_0$ is obtained by integrating
\begin{equation}
\frac{\mathrm{d}z_\tau}{\mathrm{d}\tau}
=v_\theta(z_\tau,\tau\mid I_t,s_t,l),
\qquad \tau:1\rightarrow0.
\label{eq:flow-ode}
\end{equation}

\section{Method}
\label{sec:method}

\subsection{Overview}
\label{sec:method:overview}
\begin{figure*}[!t]
    \centering
    \includegraphics[width=0.95\linewidth]{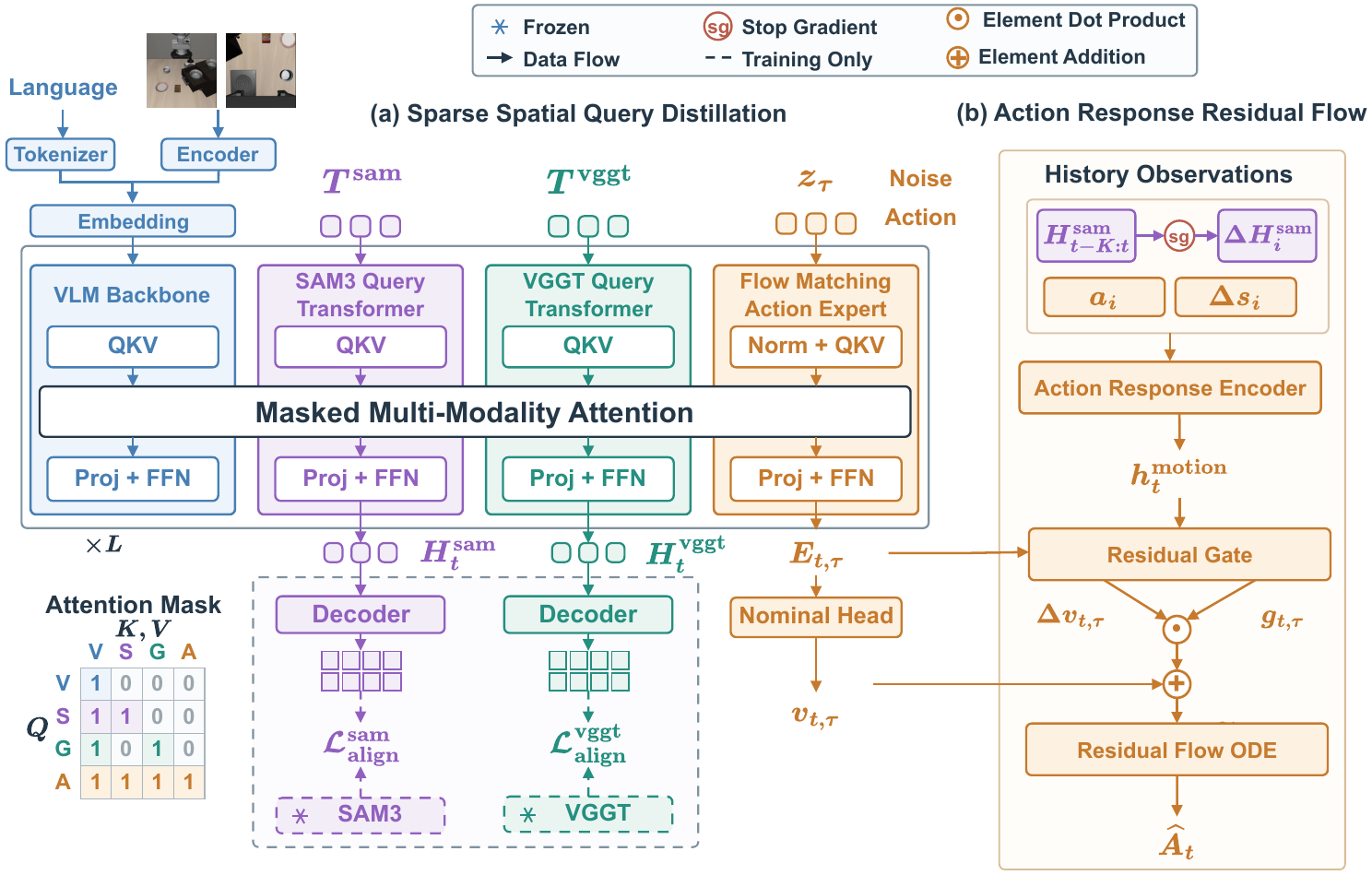}
    \caption{\textbf{SGMR Network Architecture.}
    (a) Sparse Spatial Query Distillation uses two independent transformer networks to encode the SAM3(S) and VGGT(G) learnable tokens(T). Multi-modal attention is masked: the VLM context(V) attends only to itself, each teacher attends to the VLM context and itself, and the two teachers cannot attend to each other. (b) Action(A) Response Residual Flow encodes history observations into a residual flow that refines the flow-matching velocity field.
    }
    \label{fig:method}
\end{figure*}

As shown in Figure~\ref{fig:method}, \textbf{Sparse Spatial Query Distillation} uses two separate query transformers to distill dense geometric and semantic knowledge from frozen VGGT and SAM3 teachers into compact token states, which give the action expert spatial context without a teacher-matching loss on the shared VLA tokens; the teachers and reconstruction decoders are used only during training. The \textbf{Action Response Residual Flow} relates recent executed actions to the subsequent changes in robot states and SAM3-aligned query states, and uses this execution context to refine the next action chunk through a residual on the velocity field.

\subsection{Sparse Spatial Query Distillation}
\label{sec:method:query}

To enhance spatial awareness while alleviate interference with pretrained visual-semantic priors, we introduce two independent sparse query transformers guided by frozen SAM3 and VGGT teachers. As seen in Figure~\ref{fig:method}(a), dense teacher features are reconstructed from compact token states, thereby routing spatial supervision through the query branches rather than directly constraining shared VLA representations.

\paragraph{Sparse Query Transformer.}
We index the SAM3 and VGGT branches by $k \in \{0,1\}$, respectively. Each branch maintains learnable tokens $T^k\in\mathbb{R}^{M_k\times d_k}$, where $M_k$ counts tokens across configured camera groups and $d_k$ is their hidden dimension. At step $t$, the query transformer $T_{\phi_k}$, with trainable parameters $\phi_k$, exploits the layerwise VLM context $\mathcal{C}_t$ encoded from images $I_t$ and instruction $l$:
\begin{equation}
H_t^k=T_{\phi_k}(T^k,\mathcal{C}_t),
\qquad H_t^k\in\mathbb{R}^{M_k\times d_k}.
\label{eq:spatial-query}
\end{equation}
Here, $T^k$ contains the learnable initial tokens, while $H_t^k$ contains their final states conditioned on the current observation and instruction. Following the attention mask illustrated in Figure~\ref{fig:method}(a), each query branch attends jointly to its own stream and the corresponding VLM context, while remaining isolated from the other query branch. Conversely, the VLM attends only within its original stream and the action expert accesses the layerwise keys and values from the VLM and both query branches.

\paragraph{Geometric-Semantic Reconstruction.}
To provide dense spatial supervision for the sparse query transformers, we employ two frozen teacher models: VGGT supplies geometric representations, while SAM3 provides semantic representations. For teacher $k$, let $F_t^k\in\mathbb{R}^{N_k\times c_k}$ denote the target feature grid aggregated across camera views, where $N_k$ is the number of dense teacher tokens and $c_k$ is their feature dimension. A teacher-specific decoder $D_{\psi_k}$, parameterized by $\psi_k$, reconstructs these dense targets from the final query states:
\begin{equation}
\widehat{F}_t^k=D_{\psi_k}(H_t^k),
\qquad M_k\ll N_k.
\label{eq:teacher-reconstruction}
\end{equation}
For each training example, the alignment loss $\mathcal{L}_{\mathrm{align}}^k$ minimizes the cosine distance between reconstructed features $\widehat{F}_t^k$ and teacher features $F_t^k$ over valid locations. During inference, the reconstruction decoders are removed, leaving only the query transformers to condition the policy. Decoder architecture, validity masking, and the full alignment objective are detailed in Appendix~\ref{app:query-implementation}.

\subsection{Action Response Residual Flow}
\label{sec:method:motion}
To support action adjustment under perturbations, we introduce \textbf{Action Response Residual Flow}, which complements static spatial representations with recent execution context. As shown in Figure~\ref{fig:method}(b), the module encodes executed actions together with subsequent changes in robot states and SAM3 token representations, and uses this context to condition a residual flow for action generation. The following analysis motivates the residual formulation and the use of local action--response measurements.

Let $s$ denote the robot state and $\eta$ the environment state, including object configurations, camera viewpoint, and illumination. With the instruction and network parameters fixed, image formation and query encoding define a mapping from $(s,\eta)$ to $(s,H^{\mathrm{sam}})$. Its Jacobians under nominal and nearby perturbed conditions are denoted by $J_{\mathrm{nom}}$ and $J_{\mathrm{pert}}$, which indicate the local visual--motor relationship.
\begin{proposition}[Local Residual Adaptation]
\label{prop:residual-adaptation}
For the same policy input, a perturbation in the local visual--motor relationship from $J_{\mathrm{nom}}$ to $J_{\mathrm{pert}}$ may require different actions to achieve the task. Let $\widetilde{v}^\star_{t,\tau}(J)$ denote the corresponding ideal action-generating field and define its nominal value as $v^\star_{t,\tau}=\widetilde{v}^\star_{t,\tau}(J_{\mathrm{nom}})$. As a local modeling assumption, suppose this field is differentiable with respect to $J$ at $J_{\mathrm{nom}}$. Then, as $J_{\mathrm{pert}}\to J_{\mathrm{nom}}$,
\begin{equation}
\begin{aligned}
\widetilde{v}^\star_{t,\tau}(J_{\mathrm{pert}})
={}&v^\star_{t,\tau}+D_J\widetilde{v}^\star_{t,\tau}(J_{\mathrm{nom}})[J_{\mathrm{pert}}-J_{\mathrm{nom}}]
\\ &+o(\|J_{\mathrm{pert}}-J_{\mathrm{nom}}\|).
\end{aligned}
\label{eq:residual-interpretation}
\end{equation}
Here, $D_J\widetilde{v}^\star_{t,\tau}(J_{\mathrm{nom}})$ is the derivative of the ideal velocity field with respect to $J$ at $J_{\mathrm{nom}}$. It maps the local visual--motor relationship change $J_{\mathrm{pert}}-J_{\mathrm{nom}}$ to the first-order velocity correction.
\end{proposition}
Appendix~\ref{app:action-generating field} provides the proof and scope. Since the local visual--motor relationship change is not directly available to the policy, we condition the learned correction on recent action--response measurements.

At replanning time $t$, $(s_t,\eta_t)$ denotes the current robot and environment states. We collect $K$ historical triplets $(a_i,\Delta s_i,\Delta H_i^{\mathrm{sam}})$, indexed by $i=t-K,\ldots,t-1$, where $a_i$ is the executed action,
$\Delta s_i=s_{i+1}-s_i$ and $\Delta H_i^{\mathrm{sam}}=H_{i+1}^{\mathrm{sam}}-H_i^{\mathrm{sam}}$.
The following proposition relates these measurements to the local response map under the current operating conditions.

\begin{proposition}[Local Responses]
\label{prop:action-response}
Let $f(a;s_t,\eta_t)$ denote the joint robot--environment state increment $(\Delta s,\Delta\eta)$ induced by action $a$ under the fixed current conditions $(s_t,\eta_t)$.
Motivated by local visual-servoing models~\citep{chaumette2006visual} and attention sensitivity analysis~\citep{castin2023smooth}, we assume that the joint state--query mapping has a locally Lipschitz Jacobian within a common smooth neighborhood. Under the local model and regularity assumptions in Appendix~\ref{app:local window}, nearby historical transitions satisfy
\begin{equation}
\begin{bmatrix}
\Delta s_i\\
\Delta H_i^{\mathrm{sam}}
\end{bmatrix}
=J_{\mathrm{pert}} f(a_i;s_t,\eta_t)+\rho_{i,t}.
\label{eq:local-visuomotor}
\end{equation}
Here, $\rho_{i,t}$ accounts for linearization error and deviations of historical conditions from $(s_t,\eta_t)$; its bound shrinks as transition sizes and these deviations decrease.
\end{proposition}
The derivation and error bound are provided in Appendix~\ref{app:local window}. These approximate input--output observations of the current response map are encoded by the \textbf{Action Response Encoder} into $h_t^{\mathrm{motion}}$, which conditions the \textbf{Residual Flow ODE} alongside the current action-expert states.

\paragraph{Action Response Encoder.}
The encoder processes $K$ historical triplets $(a_i,\Delta s_i,\Delta H_i^{\mathrm{sam}})$ defined above. As shown in Figure~\ref{fig:method}(b), the encoder uses each action and robot-state change to attend to the query difference, then aggregates the transition descriptors in temporal order into $h_t^{\mathrm{motion}}$. Transitions are consecutive control steps ending at the current observation, with frames before the episode start replaced by its first frame. This context is computed once per replanning step and reused during action-chunk generation.

\paragraph{Residual Flow ODE.}
Motivated by Proposition~\ref{prop:residual-adaptation}, we parameterize the corrected velocity as a nominal prediction plus a learned history-conditioned residual. At flow time $\tau$, the final action-expert states $E_{t,\tau}$ and $h_t^{\mathrm{motion}}$ jointly condition a velocity residual $\Delta v_{t,\tau}$ and a sigmoid gate $g_{t,\tau}$. The corrected field is
\begin{equation}
\widetilde{v}_{t,\tau}
=v_{t,\tau}
+
g_{t,\tau}\odot\Delta v_{t,\tau},
\label{eq:motion-residual}
\end{equation}
where $v_{t,\tau}$ is the output of a trainable head on $E_{t,\tau}$ and $\odot$ denotes element-wise multiplication. The residual and gate are recomputed at each flow step. The encoder projections, temporal aggregation, and residual heads are specified in Appendix~\ref{app:motion-implementation}.

\subsection{Training Objective}
\label{sec:method:training}

We train SGMR by jointly optimizing the corrected flow-matching objective and the two teacher-feature alignment losses:
\begin{equation}
\mathcal{L}_{\mathrm{total}}=
\mathcal{L}_{\mathrm{flow}}(\widetilde{v})
+
\lambda_{\mathrm{sam}}\mathcal{L}_{\mathrm{align}}^{\mathrm{sam}}
+
\lambda_{\mathrm{vggt}}\mathcal{L}_{\mathrm{align}}^{\mathrm{vggt}},
\label{eq:total-loss}
\end{equation}
Here, $\mathcal{L}_{\mathrm{flow}}(\widetilde{v})$ evaluates Eq.~\eqref{eq:flow} with the corrected, history-conditioned field in Eq.~\eqref{eq:motion-residual} and the coefficients $\lambda_{\mathrm{sam}}$ and $\lambda_{\mathrm{vggt}}$ weight the two alignment losses.

\section{Experiments}
\label{sec:experiments}
Our experiments aim to: 1) Assess the performance of SGMR against existing VLAs on standard benchmarks. 2) Evaluate its generalization across different distribution shifts. 3) Analyze the contribution of each component in SGMR. 4) Examine the spatial representations of the query transformers. 5) Assess the interference with pretrained VLM priors. 6) Validate the adaptive action correction capability of the Action Response Residual Flow (Appendix~\ref{app:action-adjustment}). 7) Evaluate physical task completion under nominal conditions and visual-noise, camera-viewpoint, and object-layout perturbations. 8) Quantify inference latency and parameter overhead (Appendix~\ref{app:inference-cost}).

\begin{table}[!htb]
\centering
\small\setlength{\tabcolsep}{2.6pt}\setlength{\fboxsep}{1pt}
\caption{\textbf{Average success rates (\%) comparison on LIBERO}.}
\label{tab:libero}
\begin{tabular}{@{}lccccc@{}}
\toprule
\textbf{Method}
& \textbf{Spatial}
& \textbf{Object}
& \textbf{Goal}
& \textbf{10}
& \textbf{Avg.} \\
\midrule
OpenVLA~\citep{kim2024openvla}
& 84.7 & 88.4 & 79.2 & 53.7 & 76.5 \\
OpenVLA-OFT~\citep{kim2025fine}
& 97.6 & 98.4 & 97.9 & 94.5 & 97.1 \\
CogVLA~\citep{li2026cogvla}
& 98.5 & 98.8 & 96.5 & 95.2 & 97.4 \\
$\pi_0$~\citep{black2024pi_0}
& 96.8 & 98.8 & 95.8 & 85.2 & 94.1 \\
$\pi_{0.5}$~\citep{intelligence2025pi_}
& 98.8 & 98.2 & 98.0 & 92.4 & 96.9 \\
VLA-Adapter~\citep{wang2026vla}
& 97.8 & 99.2 & 97.2 & 95.0 & 97.3 \\
MemoryVLA~\citep{shi2026memoryvla}
& 98.4 & 98.4 & 96.4 & 93.4 & 96.7 \\
Evo-1~\citep{lin2026evo}
& 92.7 & 97.7 & 96.3 & 92.3 & 94.8 \\
GeoPredict~\citep{qian2026geopredict}
& 98.0 & 98.2 & 95.7 & 94.0 & 96.5 \\
AVA-VLA~\citep{xiao2026ava}
& 97.4 & 99.4 & 97.4 & \colorbox{lightred}{97.6} & 98.0 \\
AAC ($\pi_{0.5}$)~\citep{liang2026adaptive}
& 99.1 & 99.2 & 98.0 & 95.2 & 97.9 \\
QuantVLA ($\pi_{0.5}$)~\citep{zhang2026quantvla}
& 98.5 & 98.0 & 98.0 & 96.0 & 97.6 \\
\midrule
\textbf{SGMR (Ours)}
& \colorbox{lightred}{99.8} & \colorbox{lightred}{99.6} & \colorbox{lightred}{98.3} & 96.0
& \colorbox{lightred}{98.4} \\
\bottomrule
\end{tabular}%
\end{table}

\begin{table}[!t]
\centering
\caption{\textbf{Comparison on VLABench(\%).}
Each setting reports success rate (SR) and progress score (PS). In-Dist., Category., and Common.\ denote in-distribution, cross-category, and commonsense settings.}
\label{tab:vlabench}
\footnotesize
\setlength{\tabcolsep}{0.65pt}
\setlength{\fboxsep}{0.5pt}
\renewcommand{\arraystretch}{1.15}
\resizebox{\columnwidth}{!}{%
\begin{tabular}{@{}l@{\hspace{2.5pt}}rr*{5}{@{\hspace{2.5pt}}rr}@{}}
\toprule
\textbf{Method} & \multicolumn{2}{c}{\textbf{In-Dist.}} & \multicolumn{2}{c}{\textbf{Category.}} & \multicolumn{2}{c}{\textbf{Common.}} & \multicolumn{2}{c}{\textbf{Instruction}} & \multicolumn{2}{c}{\textbf{Texture}} & \multicolumn{2}{c}{\textbf{Overall.}} \\
\cmidrule(lr){2-3}\cmidrule(lr){4-5}\cmidrule(lr){6-7}\cmidrule(lr){8-9}\cmidrule(lr){10-11}\cmidrule(lr){12-13}
 & \textbf{SR} & \textbf{PS} & \textbf{SR} & \textbf{PS} & \textbf{SR} & \textbf{PS} & \textbf{SR} & \textbf{PS} & \textbf{SR} & \textbf{PS} & \textbf{SR} & \textbf{PS} \\
\midrule
$\pi_0$ & 47.0 & 62.7 & 21.2 & 33.6 & 29.1 & 43.0 & 17.3 & 38.7 & 32.2 & 42.5 & 29.4 & 44.1 \\
$\pi_0$-FAST & 56.2 & 66.8 & 31.0 & 38.2 & 38.0 & 48.6 & 35.0 & 45.0 & 39.0 & 49.0 & 39.8 & 49.5 \\
X-VLA & -- & 67.8 & -- & 25.1 & -- & 48.2 & -- & 63.1 & -- & -- & -- & 51.1 \\
ACoT-VLA & -- & 66.1 & -- & 38.9 & -- & 37.8 & -- & 39.6 & -- & 54.6 & -- & 47.4 \\
$\pi_{0.5}$ & 65.4 & 77.8 & 38.2 & \colorbox{lightred}{49.7} & 43.9 & \colorbox{lightred}{57.3} & 48.2 & 64.2 & 44.9 & 62.3 & 48.1 & 62.3 \\
\midrule
\textbf{SGMR (Ours)} & \colorbox{lightred}{69.7} & \colorbox{lightred}{78.8} & \colorbox{lightred}{40.0} & 46.8 & \colorbox{lightred}{48.3} & 55.9 & \colorbox{lightred}{59.0} & \colorbox{lightred}{66.7} & \colorbox{lightred}{52.4} & \colorbox{lightred}{65.1} & \colorbox{lightred}{54.1} & \colorbox{lightred}{63.0} \\
\bottomrule
\end{tabular}%
}
\end{table}

\subsection{Experiments Setup}
\label{sec:experiments:setup}

\paragraph{Benchmarks and Evaluation Metrics.}
We evaluate SGMR on LIBERO~\citep{liu2023libero}, LIBERO-Plus~\citep{fei2025libero}, LIBERO-PRO~\citep{zhou2025libero} and VLABench~\citep{zhang2025vlabench}. To evaluate robustness, we transfer the LIBERO-trained model to LIBERO-Plus and LIBERO-PRO without additional fine-tuning. LIBERO-Plus covers seven perturbation dimensions: camera viewpoint, robot initial state, language instruction, lighting, background, visual noise, and scene layout; LIBERO-PRO covers object, position, semantic, and task perturbations. VLABench evaluates in-distribution, cross-category, commonsense, semantic-instruction, and unseen-texture settings. We report success rate (SR, \%) on all benchmarks and progress score (PS, \%) on VLABench.

\paragraph{Experiments Setting.}
We use OpenPI as the training framework and $\pi_{0.5}$~\citep{intelligence2025pi_} as the baseline model. Training is conducted on 8 H20 GPUs for all datasets, with 128 total batch size and the AdamW optimizer~\citep{loshchilov2017decoupled}. For the dual-teacher distillation transformers, learnable tokens are organized per input camera. Both the SAM3 and VGGT teachers employ the main and wrist cameras to supervise the reconstruction of sparse geometric and semantic features, with loss weights of 0.4 and 0.1, respectively. For each camera, we use 16 learnable tokens, each of dimension 512. More experiment settings are in Table~\ref{tab:training-model-configuration} and details about baselines are in Appendix~\ref{app:compared-models}.

\subsection{Main results}
\label{sec:experiments:main}

On LIBERO (Table~\ref{tab:libero}), SGMR achieves the highest average success rate (98.4\%) and improves on the reported $\pi_{0.5}$ in all four suites, most on LIBERO-10 (92.4\% to 96.0\%). On VLABench (Table~\ref{tab:vlabench}), it raises the in-distribution SR from 65.4\% to 69.7\% and the overall SR from 48.1\% to 54.1\%.

\subsection{Generalization across distribution shifts}
We transfer the LIBERO-trained policy to LIBERO-Plus and LIBERO-PRO without fine-tuning. On LIBERO-Plus (Table~\ref{tab:libero-plus}), SGMR reaches 87.6\%, 3.0 points above the released $\pi_{0.5}$, with the largest gain under robot initial-state perturbations (75.5\% to 86.6\%) and further gains under noise and camera shifts. On LIBERO-PRO (Table~\ref{tab:libero-pro}), SGMR raises overall success from 59.2\% to 62.2\% and improves 11 of the 16 suite--perturbation combinations, including semantic perturbations in all four suites and position and task perturbations in three. On VLABench (Table~\ref{tab:vlabench}), SGMR improves SR over $\pi_{0.5}$ in all four out-of-distribution settings, most under instruction shifts (48.2\% to 59.0\%), whereas the average progress score rises only from 62.3\% to 63.0\% and falls in the cross-category and commonsense settings.

\begin{table}[!htb]
\centering
\caption{\textbf{Zero-Shot success rates (\%) comparison on LIBERO-Plus}. Cam., Rob., Lang., and BG. denote camera viewpoint, robot initial state,language instruction, and background perturbations.
}
\label{tab:libero-plus}
\scriptsize
\setlength{\tabcolsep}{3pt}
\renewcommand{\arraystretch}{1.10}
\resizebox{\linewidth}{!}{%
\begin{tabular}{@{}l*{8}{c}@{}}
\toprule
\textbf{Method}
& \textbf{Cam.}
& \textbf{Rob.}
& \textbf{Lang.}
& \textbf{Light}
& \textbf{BG.}
& \textbf{Noise}
& \textbf{Layout}
& \textbf{Overall} \\
\midrule
OpenVLA-OFT~\citep{kim2025fine}
& 56.4 & 31.9 & 79.5 & 88.7 & 93.3 & 75.8 & 74.2 & 69.6 \\
OpenVLA~\citep{kim2024openvla}
& 0.8 & 3.5 & 23.0 & 8.1 & 34.8 & 15.2 & 28.5 & 15.6 \\
SRPO~\citep{fei2026srpo}
& 17.1 & 51.0 & 81.8 & 70.4 & 88.9 & 35.3 & 72.4 & 59.6 \\
MergeVLA~\citep{fu2026mergevla}
& 64.4 & 46.4 & 75.7 & 92.7 & 92.7 & 74.7 & 73.7 & 72.4 \\
VLA-JEPA~\citep{sun2026vla}
& 63.3 & 67.1 & 85.4 & 95.6 & 93.6 & 66.3 & 85.1 & 79.5 \\
Spatial Forcing~\citep{li2026spatial}
& 0.1 & 0.3 & 26.8 & 66.0 & 45.9 & 0.1 & 59.8 & 25.7 \\
$\pi_0$~\citep{black2024pi_0}
& 61.0 & 40.8 & 63.5 & 89.3 & 84.1 & 80.1 & 76.4 & 69.4 \\
$\pi_0$-FAST~\citep{pertsch2025fast}
& 65.1 & 21.6 & 61.0 & 73.2 & 73.2 & 74.4 & 68.8 & 61.6 \\
RIPT-VLA~\citep{tan2025interactive}
& 55.2 & 31.2 & 77.6 & 88.4 & 91.6 & 73.5 & 74.2 & 68.4 \\
$\pi_{0.5}$~\citep{intelligence2025pi_}
& 71.7 & 75.5 & \colorbox{lightred}{85.9}
& \colorbox{lightred}{96.1} & \colorbox{lightred}{95.7}
& 86.4 & 87.5 & 84.6 \\
\midrule
\textbf{SGMR (Ours)}
& \colorbox{lightred}{75.0} & \colorbox{lightred}{86.6} & \colorbox{lightred}{85.9}
& 94.8 & 94.1 & \colorbox{lightred}{93.0}
& \colorbox{lightred}{87.9} & \colorbox{lightred}{87.6} \\
\bottomrule
\end{tabular}%
}
\end{table}

\begin{table}[!t]
\centering
\caption{\textbf{Zero-Shot success rates (\%) comparison on LIBERO-PRO.} Obj., Pos., and Sem.\ denote object, position, and semantic perturbations.}
\label{tab:libero-pro}
\small
\setlength{\tabcolsep}{1pt}
\setlength{\fboxsep}{1pt}
\renewcommand{\arraystretch}{1.05}
\begin{tabular*}{\columnwidth}{@{\extracolsep{\fill}}l*{8}{r}@{}}
\toprule
\textbf{Method} & \multicolumn{4}{c}{\textbf{LIBERO-Spatial}} & \multicolumn{4}{c}{\textbf{LIBERO-Object}} \\
\cmidrule(lr){2-5}\cmidrule(lr){6-9}
 & \textbf{Obj.} & \textbf{Pos.} & \textbf{Sem.} & \textbf{Task} & \textbf{Obj.} & \textbf{Pos.} & \textbf{Sem.} & \textbf{Task} \\
\midrule
$\pi_0$ & 95.4 & 8.0 & 88.4 & \colorbox{lightred}{55.0} & 87.0 & 0.0 & 98.6 & 9.8 \\
$\pi_{0.5}$ & \colorbox{lightred}{97.4} & \colorbox{lightred}{49.6} & 97.0 & 54.4 & 94.0 & 18.2 & 99.0 & 10.4 \\
\midrule
\textbf{SGMR} & 97.0 & 49.4 & \colorbox{lightred}{99.4} & 52.0 & \colorbox{lightred}{96.8} & \colorbox{lightred}{44.8} & \colorbox{lightred}{99.4} & \colorbox{lightred}{11.6} \\
\midrule
\end{tabular*}
\par\smallskip
\begin{tabular*}{\columnwidth}{@{\extracolsep{\fill}}l*{9}{r}@{}}
\textbf{Method} & \multicolumn{4}{c}{\textbf{LIBERO-Goal}} & \multicolumn{4}{c}{\textbf{LIBERO-10}} & \textbf{All suites} \\
\cmidrule(lr){2-5}\cmidrule(lr){6-9}\cmidrule(l){10-10}
 & \textbf{Obj.} & \textbf{Pos.} & \textbf{Sem.} & \textbf{Task} & \textbf{Obj.} & \textbf{Pos.} & \textbf{Sem.} & \textbf{Task} & \textbf{Avg.} \\
\midrule
$\pi_0$ & 71.0 & 9.6 & 82.6 & 10.6 & 44.0 & 0.0 & 72.8 & 9.0 & 46.4 \\
$\pi_{0.5}$ & \colorbox{lightred}{88.8} & 34.8 & 96.4 & 20.6 & \colorbox{lightred}{67.2} & 9.6 & 92.4 & 17.8 & 59.2 \\
\midrule
\textbf{SGMR} & 86.0 & \colorbox{lightred}{37.8} & \colorbox{lightred}{97.8} & \colorbox{lightred}{21.2} & 66.2 & \colorbox{lightred}{15.6} & \colorbox{lightred}{96.2} & \colorbox{lightred}{25.8} & \colorbox{lightred}{62.2} \\
\bottomrule
\end{tabular*}
\end{table}

\subsection{Ablation Studies}
\label{sec:experiments:ablation}

\begin{figure*}[!t]
\centering
\includegraphics[width=0.98\textwidth]{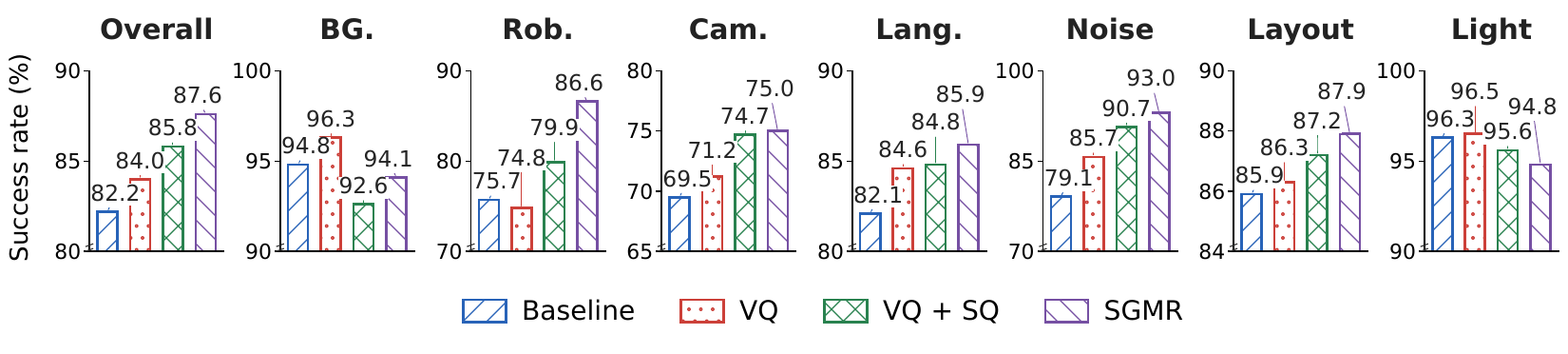}
\caption{\textbf{Component Ablations on LIBERO-Plus Perturbations.}
VQ/SQ denote VGGT/SAM3 query transformer.}
\label{fig:ablation-module}
\end{figure*}

\paragraph{Component of SGMR.} Figure~\ref{fig:ablation-module} adds the components cumulatively on LIBERO-Plus to $\pi_{0.5}$ retrained with the SGMR recipe (Table~\ref{tab:training-model-configuration}). This baseline differs from the released $\pi_{0.5}$ checkpoint in training, with a global batch size of 128 instead of 256 and normalization statistics recomputed on our LIBERO copy. The VGGT and SAM3 query branches raise overall success to 84.0\% and 85.8\%, and the residual flow to 87.6\%, 5.4 points above the retrained baseline, with the largest gains under noise and robot initial-state perturbations.

\begin{table}[!t]
\centering
\small
\setlength{\tabcolsep}{2.5pt}\setlength{\fboxsep}{1pt}
\caption{\textbf{Structure and input ablations on LIBERO-Plus.} SR (\%) Cam. and Rob. perturbations and Overall.}
\label{tab:history-supervision}
\begin{tabular}{@{}lrrr@{}}
\toprule
Configuration & Cam. & Rob. & Overall \\
\midrule
\multicolumn{4}{@{}l}{\emph{Spatial query supervision}} \\
$\pi_{0.5}$ & 69.5 & 75.7 & 82.2 \\
Dual queries, no reconstruction & 71.4 & 73.5 & 83.2 \\
Dual queries, with reconstruction & \colorbox{lightred}{74.7} & \colorbox{lightred}{79.9} & \colorbox{lightred}{85.8} \\
\midrule
\multicolumn{4}{@{}l}{\emph{Temporal conditioning}} \\
Standard history conditioning & 68.4 & 84.8 & 85.4 \\
Residual flow, action + proprio. & 72.0 & 86.0 & 85.6 \\
Residual flow, visual + proprio. & 72.0 & 78.1 & 85.2 \\
Residual flow, action + visual. & 72.5 & 86.3 & 87.2 \\
Residual flow, shuffled action response. & 71.7 & 84.8 & 86.0 \\
\textbf{SGMR} & \colorbox{lightred}{75.0} & \colorbox{lightred}{86.6} & \colorbox{lightred}{87.6} \\
\bottomrule
\end{tabular}

\end{table}

\paragraph{Structure and input.}
Table~\ref{tab:history-supervision} separates the modules from added capacity and input. With the same dual-query architecture, reconstruction raises SR from 83.2\% to 85.8\%, while the queries alone add 1.0 point, so the spatial gain stems from teacher supervision. For temporal conditioning, history alone is not enough. Standard history conditioning, which feeds $h_t^{\mathrm{motion}}$ to the action expert as an extra input, gains under robot initial-state changes but loses under camera shifts, and reaches 85.4\% overall, which demonstrates the effectiveness of the Residual Flow ODE architecture. Among the residual-flow variants, SGMR, which uses complete triplets with each action paired with its own response, performs best overall and under perturbations, ahead of the variants without actions, without visual responses and without proprio. This agrees with Proposition~\ref{prop:action-response}: correctly paired actions and responses measure the current visual--motor relationship, which the residual flow uses to correct the next actions under perturbations. Moreover, residual-flow variants with shuffled disrupts triplet correspondences and impairs adaptation under Cam. and Rob. perturbations, suggesting that our gains stem from visual–motor relationships rather than history alone.

\subsection{Spatial Representation Analysis}
\label{sec:experiments:query-representations}

\begin{figure}[!t]
\centering
\includegraphics[width=\linewidth]{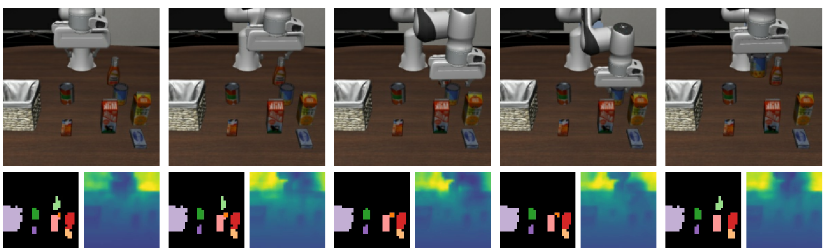}
\caption{\textbf{Linear probes on frozen query tokens.} Top: main-camera frames. Bottom: predicted body IDs (left) and depth (right) for each frame.}
\label{fig:query-spatial-probes}
\end{figure}

To check whether the query tokens carry spatial information, we freeze the policies and fit linear probes on the final query tokens of separately trained SAM3- and VGGT-query policies: an affine probe that maps each 512-dimensional SAM3 token to $32\times32$ body-ID logits, and a linear depth probe on the VGGT tokens. Figure~\ref{fig:query-spatial-probes} shows the predictions on main-camera frames 0, 20, 40, 60 and 90 of LIBERO demonstration episode~8 (``put both the alphabet soup and the tomato sauce in the basket''). The segmentation probe separates the basket and the tabletop objects and the depth probe follows the coarse geometry around the moving arm, so the query tokens encode both the semantic layout and the geometry of the scene. The body IDs are object instances of the scene and the depth is shown in relative relation.

\subsection{Interference with Pretrained VLM Priors}
As all models are post-trained on robot demonstrations without visual question answering (VQA) data. To assess interference with pretrained VLM priors, we (i) construct a language-extended benchmark based on LIBERO-Plus to evaluate instruction following and (ii) measure centered kernel alignment (CKA) similarity between VLM representations before and after post-training to assess representational drift.
\label{sec:experiments:prior}

\begin{table}[!t]
\centering
\caption{\textbf{VLM Interference Analysis.} SR (\%) under the original instructions (Orig.), semantic rewrites (lexical, commonsense, and multilingual; Sem.), and LIBERO-Plus rewrites (LP); CKA to the pretrained $\pi_{0.5}$ base, averaged over layers.}
\label{tab:language-prior}
\footnotesize
\setlength{\tabcolsep}{2pt}\setlength{\fboxsep}{1pt}
\resizebox{\columnwidth}{!}{%
\begin{tabular}{@{}lccc@{\hspace{6pt}}ccc@{\hspace{6pt}}ccc@{}}
\toprule
 & \multicolumn{3}{c}{Split A SR (\%)} & \multicolumn{3}{c}{Split B SR (\%)} & \multicolumn{3}{c}{CKA to $\pi_{0.5}$ base} \\
\cmidrule(lr){2-4}\cmidrule(lr){5-7}\cmidrule(l){8-10}
Model & Orig. & Sem. & LP & Orig. & Sem. & LP & Main & Wrist & Lang. \\
\midrule
$\pi_{0.5}$ & 98.3 & 58.4 & 65.1 & 95.2 & 85.1 & 89.1 & 0.43 & 0.40 & 0.67 \\
VGGT align & 95.8 & 54.7 & 69.0 & 97.6 & 78.0 & 84.1 & 0.39 & 0.34 & 0.63 \\
SAM3 align & 97.5 & 53.3 & 62.3 & 97.6 & 70.3 & 77.6 & 0.27 & 0.23 & 0.64 \\
VQ & 95.8 & 59.7 & \colorbox{lightred}{76.8} & 97.6 & 83.2 & 89.1 & \colorbox{lightred}{0.51} & 0.43 & 0.70 \\
SQ & 100.0 & 60.2 & 75.2 & 97.6 & 86.5 & 91.1 & 0.45 & 0.37 & 0.69 \\
\midrule
SGMR & 99.2 & \colorbox{lightred}{60.3} & 76.4 & 97.6 & \colorbox{lightred}{86.9} & \colorbox{lightred}{92.4} & 0.45 &\colorbox{lightred}{0.43} & \colorbox{lightred}{0.72} \\
\bottomrule
\end{tabular}
}
\end{table}

\paragraph{Language-extended Benchmark.}
Direct alignment adds a teacher-matching loss inside the VLM, which also encodes the instruction. To test whether this hurts instruction following, we extend the language tests of LIBERO-Plus~\citep{fei2025libero} on its 40 tasks with more rewrites, control instructions and initial states (6{,}238 episodes per model, Appendix~\ref{app:language-benchmark}). In Split~A, 12 tasks share two scenes, so only the instruction tells the policy what to do: without it, success falls below 20\%. The other 28 tasks form Split~B. We compare $\pi_{0.5}$, its two direct-alignment variants (VGGT/SAM3 align), the single-branch query models VQ and SQ, and SGMR. Table~\ref{tab:language-prior} reports success with the original training instruction (Orig.), with semantic rewrites that use synonyms, describe objects by commonsense or translate the instruction into Chinese, Spanish or French (Sem.), and with all LIBERO-Plus rewrites of the task (LP).

\paragraph{Instruction following.}
Direct alignment clearly affect robustness to rewritten instructions, while query distillation does not (Table~\ref{tab:language-prior}). In Split~B, VGGT and SAM3 alignment lose 9.5 and 17.1 points more than $\pi_{0.5}$ on semantic rewrites and 7.4 and 13.9 points more on LP rewrites, whereas VQ, SQ and SGMR never lose significantly more than $\pi_{0.5}$. Split~A is harder for all models, which lose about 40 points on semantic rewrites. On its LP rewrites, VQ and SGMR even lose 14.3 and 10.5 points less than $\pi_{0.5}$.

\paragraph{Representation drift.}
Query distillation also changes the pretrained VLM less. In linear CKA to the pretrained $\pi_{0.5}$ base, all models drift considerably, but VQ, SQ and SGMR stay closer to the base than both alignment variants on main-camera, wrist-camera and language tokens, and closer than $\pi_{0.5}$ on main-camera and language tokens (more results in Appendix~\ref{app:language-benchmark}).
Together, the results indicate that routing teacher supervision through query transformers interferes less with the pretrained VLM than aligning the VLM features directly.

\subsection{Real-World Experiments}
\label{sec:experiments:real-world}

\paragraph{Setup and tasks.}
We evaluate SGMR and $\pi_{0.5}$ on a six-degree-of-freedom Alicia-D
arm with a parallel gripper and main/wrist RGB cameras. The 150 nominal
demonstrations cover block stacking (50), block-to-plate placement (50),
and earphones-to-box placement (50), with fixed instructions per task.
Besides nominal conditions, we change the main-camera pitch over
$[90^{\circ}, 120^{\circ}]$ and yaw over $[-60^{\circ}, 60^{\circ}]$, corrupt the
main-camera images with 15 frozen noise configurations from five LIBERO-Plus
families~\citep{fei2025libero}, identical for both models, and rearrange
objects and receptacles. Appendix~\ref{app:real-world-recordings} details experiment settings.

\begin{figure}[!t]
  \centering
  \includegraphics[width=\linewidth]{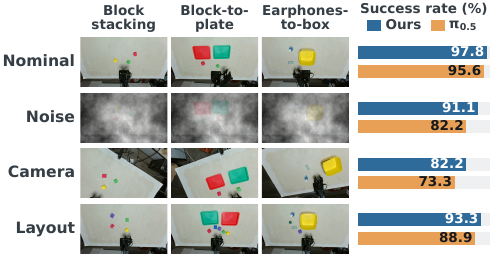}
  \caption{\textbf{Real-world tasks, perturbation examples, and success rates.}}
  \label{fig:real-world-overview}
\end{figure}

\begin{figure}[!t]
  \centering
  \includegraphics[width=\linewidth]{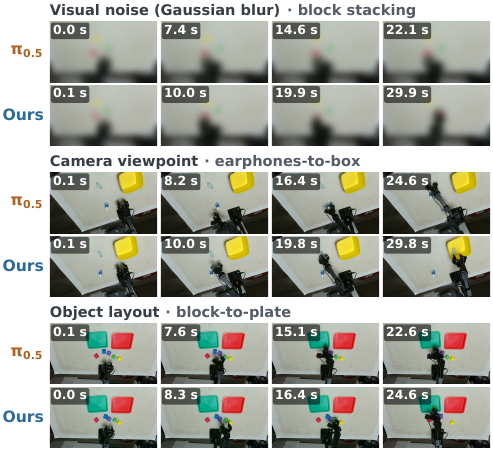}
  \caption{\textbf{Recorded real-world executions.}}
  \label{fig:real-world-executions}
\end{figure}

\paragraph{Results.}
As seen in Figure~\ref{fig:real-world-overview}, SGMR succeeds more often than $\pi_{0.5}$ in every condition: 97.8\% vs.\ 95.6\% under nominal conditions, and 8.9, 8.9 and 4.4 points more under noise, viewpoint and layout perturbation. As on LIBERO-Plus, the gains under noise and viewpoint perturbation exceed that under layout changes, indicating that its robustness also holds on a real robot. Some real world executions can be seen in Figure~\ref{fig:real-world-executions}.

\section{Conclusion}
\label{sec:conclusion}

SGMR couples \textbf{Sparse Spatial Query Distillation} with an \textbf{Action Response Residual Flow} to enhance spatial representations and adapt action generation from recent execution context. Experiments on LIBERO, VLABench, LIBERO-Plus, and LIBERO-PRO demonstrate outstanding manipulation performance and improved zero-shot robustness, with ablations confirming complementary gains from the spatial and temporal components.

\FloatBarrier
\section*{LLM Usage Statement}
We used OpenAI Codex to format the manuscript, update the method name, and prepare the revised real-world and ablation tables and descriptions from the recorded results.

  \maketitlesupplementary
  \pdfbookmark[0]{Supplementary Material}{supplementary-material}
  \backrefsetup{disable}
\appendix
\setcounter{figure}{0}
\setcounter{table}{0}
\setcounter{equation}{0}
\renewcommand{\thefigure}{S\arabic{figure}}
\renewcommand{\thetable}{S\arabic{table}}
\renewcommand{\theequation}{S\arabic{equation}}
\renewcommand{\theHfigure}{S\arabic{figure}}
\renewcommand{\theHtable}{S\arabic{table}}
\renewcommand{\theHequation}{S\arabic{equation}}

\section{Local Visual--Motor Analysis of Action Response Residual Flow}
\label{app:local-arc}

This appendix provides proofs for Propositions~\ref{prop:residual-adaptation} and~\ref{prop:action-response}, together with additional interpretations of the Action Response Residual Flow. These analyses serve only as theoretical motivation and do not introduce any additional estimator, controller, or training objective. In implementation, the module directly encodes observed action--response transitions and predicts a residual correction to the generative velocity field, without explicitly estimating the local Jacobians.

\subsection{Physical Motion and Visual Changes}
\label{app:local-model}

For a fixed camera, a three-dimensional point
$P_c=(X_c,Y_c,D_c)$ with positive depth projects to
\begin{equation}
\begin{aligned}
\pi(P_c)&=
\begin{bmatrix}
f_xX_c/D_c+c_x\\ f_yY_c/D_c+c_y
\end{bmatrix},\\[2pt]
D\pi(P_c)&=
\begin{bmatrix}
f_x/D_c&0&-f_xX_c/D_c^2\\
0&f_y/D_c&-f_yY_c/D_c^2
\end{bmatrix}.
\end{aligned}
\label{eq:app-projection}
\end{equation}
Here $f_x,f_y,c_x,c_y$ are intrinsic camera parameters. If the point
is attached to a robot link, let $q^{\mathrm{rob}}$ be local robot
configuration coordinates, $J_{\mathrm{kin}}$ its position Jacobian,
and $R_{cw}$ the world-to-camera rotation. The chain rule gives
\begin{equation}
\Delta\pi
=D\pi(P_c)R_{cw}J_{\mathrm{kin}}\Delta q^{\mathrm{rob}}
+\mathcal O(\|\Delta q^{\mathrm{rob}}\|^2).
\label{eq:app-image-motion}
\end{equation}
This recovers the local interaction-matrix perspective commonly used in visual servoing~\citep{chaumette2006visual}: the mapping from physical motion to image-space displacement depends explicitly on depth, viewpoint, and robot configuration. Since the projection of a single point has rank at most two, its image coordinates alone do not determine a unique three-dimensional motion. This derivation is intended only as an illustrative local example. Our latent query representations are not tracked image points, and practical manipulation additionally involves moving wrist cameras, object motion, and contact dynamics. We therefore adopt the more general local state representation introduced below.

Fix the language instruction and the network parameters. Let $\xi=(s,\eta)$ collect the robot state $s$ and the environment state $\eta$ needed to describe the local configuration and its transitions, including object configurations, camera viewpoint, and illumination. Commas separate components of this ordered pair. In the noiseless local model, the multi-view RGB images $I$ are observations formed under these conditions; the fixed vision-language and query encoders then produce $H^{\mathrm{sam}}$ from $I$ and the instruction. Thus, the joint map below includes both image formation and query encoding. The images are intermediate observations of the underlying robot and environment states, not additional coordinates of $\xi$. We write
\[
\Phi(\xi)=
\begin{bmatrix}
s\\
H^{\mathrm{sam}}
\end{bmatrix}.
\]
Brackets collect the robot-state vector and query-token matrix as two blocks. The norm on this product space combines the Euclidean norm of the state block and the Frobenius norm of the query block in quadrature. The Jacobian acts as a linear map into this product space, so no explicit flattening is needed. Assume that $\Phi$ is continuously differentiable on a convex neighborhood containing the nominal and current operating points and all historical transition segments considered below, and that its Jacobian
\[
J(\xi)=\frac{\partial \Phi(\xi)}{\partial \xi}
\]
is $L_J$-Lipschitz in operator norm. For the perturbed rollout, let $t$ denote the current replanning time and $\xi_t=(s_t,\eta_t)$ the robot and environment states at that time. The corresponding Jacobian is $J_{\mathrm{pert}}=J(\xi_t)$: $t$ indexes execution time, while $\mathrm{pert}$ labels the operating conditions.

\paragraph{Local regularity.}
Visual servoing relates visual-feature changes to physical motion through interaction matrices~\citep{chaumette2006visual}, while \citet{castin2023smooth} bound self-attention sensitivity on compact token domains. These results motivate a local analysis; they do not establish the Lipschitz-Jacobian condition for the full state--query map. A sufficient condition is that observation formation and the fixed query encoder are twice continuously differentiable on an open neighborhood with unchanged visibility and contact modes. With a fixed attention mask, their composition then has bounded second derivatives on a compact convex subset, yielding a Lipschitz Jacobian there. The state block is linear. We retain this as a local modeling assumption, with noise and changes of visibility or contact regime discussed below.

\subsection{How Perturbations Change the Local Map}

Under a fixed instruction and fixed network parameters, let
$\xi_t^{\mathrm{nom}}=(s_t^{\mathrm{nom}},\eta_t^{\mathrm{nom}})$ denote a nominal operating point and
$\xi_t=\xi_t^{\mathrm{nom}}+e_t$ a nearby perturbed operating point, with both lying within the same smooth neighborhood. The perturbation $e_t$ compares nominal and perturbed operating conditions at the current replanning step; it does not denote the state increment of a historical transition. The labels $\mathrm{nom}$ and $\mathrm{pert}$ distinguish operating conditions at the same replanning time $t$, rather than two time steps. In particular, $\xi_t^{\mathrm{nom}}$ is a nearby reference at time $t$, not the initial state of the rollout. Writing $J_{\mathrm{nom}}=J(\xi_t^{\mathrm{nom}})$ and $\delta J=J_{\mathrm{pert}}-J_{\mathrm{nom}}$, the $L_J$-Lipschitz continuity of $J$ implies
\begin{equation}
\|\delta J\|
=\|J(\xi_t^{\mathrm{nom}}+e_t)-J(\xi_t^{\mathrm{nom}})\|
\le L_J\|e_t\|.
\label{eq:app-perturbation-map-bound}
\end{equation}
Thus, $\delta J$ characterizes the change in the local representation map induced by the perturbation $e_t$, rather than the physical perturbation itself. Equation~\eqref{eq:app-image-motion} provides a concrete geometric interpretation: viewpoint and depth modify the projection geometry and camera transformation, whereas robot configuration alters the kinematic Jacobian and the corresponding point location. This local analysis applies to continuous changes in operating conditions; it does not treat instruction variations or arbitrary image noise as smooth perturbations of the same local map.

\subsection{Residual Adaptation of the Action-Generating Field}
\label{app:action-generating field}

Even for the same policy input, a change in the local visual--motor relationship may require different actions to achieve the task. We describe this dependence through the ideal velocity field of the action-generating ODE. Here, $J$ denotes the joint robot-state and query Jacobian defined above, with $J_{\mathrm{nom}}=J(\xi_t^{\mathrm{nom}})$ serving as the nominal local map. We write $\widetilde{v}^\star_{t,\tau}(J)$ for the ideal field at fixed current policy inputs, including the noisy action chunk $z_\tau$, flow time $\tau$, and conditioning context, and define its nominal value as $v^\star_{t,\tau}=\widetilde{v}^\star_{t,\tau}(J_{\mathrm{nom}})$. These symbols mirror the nominal and corrected learned velocities $v_{t,\tau}$ and $\widetilde{v}_{t,\tau}$, while $\star$ distinguishes the ideal fields from their learned counterparts.
\begin{proof}[Proof of Proposition~\ref{prop:residual-adaptation}]
Equation~\eqref{eq:app-perturbation-map-bound} gives
$\|\delta J\|\le L_J\|e_t\|$.
As a local modeling assumption, $\widetilde{v}^\star_{t,\tau}(J)$ is Fr\'echet differentiable with respect to $J$ at $J_{\mathrm{nom}}$, with the current policy inputs held fixed. By the definition of differentiability,

\begin{equation}
\widetilde{v}^\star_{t,\tau}(J_{\mathrm{nom}}+\Delta J)
=v^\star_{t,\tau}
+D_J\widetilde{v}^\star_{t,\tau}(J_{\mathrm{nom}})[\Delta J]
+R(\Delta J),
\label{eq:app-flow-residual}
\end{equation}
Here, $D_J\widetilde{v}^\star_{t,\tau}(J_{\mathrm{nom}})$ is the derivative of the ideal velocity field with respect to $J$, evaluated at $J_{\mathrm{nom}}$ while holding the current policy inputs fixed. It maps a Jacobian change $\Delta J$ to the corresponding first-order change in the ideal velocity field.
The remainder $R$ satisfies
\begin{equation}
R(0)=0,\qquad
\lim_{\substack{\Delta J\to 0\\\Delta J\ne 0}}
\frac{\|R(\Delta J)\|}{\|\Delta J\|}=0.
\label{eq:app-flow-remainder-bound}
\end{equation}
Setting $\Delta J=\delta J$ gives the $o(\|J_{\mathrm{pert}}-J_{\mathrm{nom}}\|)$ remainder in Eq.~\eqref{eq:residual-interpretation}, proving Proposition~\ref{prop:residual-adaptation}. Equation~\eqref{eq:app-perturbation-map-bound} ensures that $\delta J\to 0$ as the robot and environment state perturbation $e_t$ tends to zero; no quadratic rate for the velocity remainder is assumed.
\end{proof}

The field $\widetilde{v}^\star_{t,\tau}(J)$ is a generative velocity over action chunks. Its differentiability with respect to $J$ is an independent local modeling assumption; it is not implied by encoder smoothness, the flow-matching objective, or the local response relation in Proposition~\ref{prop:action-response}. In practice, a perturbation may simultaneously affect the current policy inputs, the local representation map $J$, and the physical transition dynamics. Our expansion isolates only the variation induced through $J$ while holding the current inputs fixed, and therefore does not capture all possible changes in the action-to-state dynamics.

\subsection{Why Differences and a Local Window}
\label{app:local window}

Proposition~\ref{prop:action-response} rests on two local assumptions: the local model of Appendix~\ref{app:local-model}, in which $\Phi$ is continuously differentiable on a common convex neighborhood that contains the current operating point and all historical transition segments, with an $L_J$-Lipschitz Jacobian $J$, and the transition regularity in Eq.~\eqref{eq:app-transition-regularity} below.

To provide observed execution context for the learned residual, we use $i=t-K,\ldots,t-1$ to index the completed historical transitions before the current replanning time $t$. In the local state coordinates defined above, let the executed command $a_i$ induce
\begin{equation}
\xi_{i+1}=\xi_i+f(a_i;s_i,\eta_i).
\label{eq:app-action-transition}
\end{equation}
Here, $f$ is a fixed, possibly nonlinear transition-increment function. The semicolon separates the action from the initial conditions under which its response is evaluated. Its output contains both robot and environment increments: $\Delta\xi_i=(\Delta s_i,\Delta\eta_i)=f(a_i;s_i,\eta_i)$, where $\Delta\eta_i=\eta_{i+1}-\eta_i$. The environment increment belongs to the theoretical state model; the encoder instead observes the visual response through $\Delta H_i^{\mathrm{sam}}$. For the short-window approximation, we use the current conditions $(s_t,\eta_t)$ as a common reference and account for deviations of historical conditions through an error term. This does not require the historical states to be identical or the state to remain unchanged during execution.
Importantly, the commanded action need not coincide with the realized state change; $a_i$ may represent a target position or another control input whose physical effect depends on the current local state. This formulation assumes only that $\xi_i$ contains the variables required to characterize the local transition. The implementation does not estimate or evaluate $f$ explicitly, but instead directly uses the executed actions and their observed responses.

\paragraph{Transition regularity.}
We additionally assume that the transition function is defined at both the historical and current initial conditions for each historical action, within this common regime, and satisfies
\begin{equation}
\|f(a_i;s_i,\eta_i)-f(a_i;s_t,\eta_t)\|
\le L_f\|\xi_i-\xi_t\|.
\label{eq:app-transition-regularity}
\end{equation}
Here, $L_f$ is a local Lipschitz constant uniform over the historical actions. This bounds the error from using the current conditions as a reference; it imposes no linearity assumption on the dependence on action. This is an independent local dynamics assumption. A sufficient condition is that $f$ is continuously differentiable in $(s,\eta)$, with derivatives uniformly bounded over the historical actions on the common convex neighborhood.

\begin{proof}[Proof of Proposition~\ref{prop:action-response}]
Using the common representation map and regularity assumptions above, denote $J_i=J(\xi_i)$ and $\Delta\xi_i=\xi_{i+1}-\xi_i$. The fundamental theorem of calculus yields
\begin{equation}
\begin{aligned}
\begin{bmatrix}
\Delta s_i\\
\Delta H_i^{\mathrm{sam}}
\end{bmatrix}
-J_{\mathrm{pert}}\Delta\xi_i
={}&(J_i-J_{\mathrm{pert}})\Delta\xi_i\\
&+\int_0^1
\big[J(\xi_i+u\Delta\xi_i)-J_i\big]
\Delta\xi_i\,\mathrm du.
\end{aligned}
\label{eq:app-local-remainder}
\end{equation}
Bounding the first term by
$L_J\|\xi_i-\xi_t\|\|\Delta\xi_i\|$ and the integral by
$\int_0^1 L_J u\|\Delta\xi_i\|^2\,\mathrm du$ yields
\begin{equation}
\|r^{\Phi}_{i,t}\|
\le L_J\|\xi_i-\xi_t\|\,\|\Delta\xi_i\|
+\frac{L_J}{2}\|\Delta\xi_i\|^2,
\label{eq:app-representation-bound}
\end{equation}
where $r^{\Phi}_{i,t}=\Phi(\xi_{i+1})-\Phi(\xi_i)-J_{\mathrm{pert}}\Delta\xi_i$ is the representation linearization remainder. Substituting $\Delta\xi_i=f(a_i;s_i,\eta_i)$ and adding and subtracting $J_{\mathrm{pert}} f(a_i;s_t,\eta_t)$ gives
\begin{equation}
\begin{aligned}
\begin{bmatrix}
\Delta s_i\\
\Delta H_i^{\mathrm{sam}}
\end{bmatrix}
&=J_{\mathrm{pert}} f(a_i;s_t,\eta_t)+\rho_{i,t},\\
\rho_{i,t}
&=r^{\Phi}_{i,t}
+J_{\mathrm{pert}}\big[f(a_i;s_i,\eta_i)-f(a_i;s_t,\eta_t)\big].
\end{aligned}
\label{eq:app-conditioned-response}
\end{equation}
Combining Eqs.~\eqref{eq:app-transition-regularity} and~\eqref{eq:app-representation-bound} yields
\begin{equation}
\begin{aligned}
\|\rho_{i,t}\|
\le{}&\big(\|J_{\mathrm{pert}}\|L_f+L_J\|\Delta\xi_i\|\big)
\|\xi_i-\xi_t\|\\
&+\frac{L_J}{2}\|\Delta\xi_i\|^2.
\end{aligned}
\label{eq:local-history-bound}
\end{equation}
This proves Proposition~\ref{prop:action-response}. The robot-state block of $r^{\Phi}_{i,t}$ is zero because that block of $\Phi$ is the identity on $s$. The total error $\rho_{i,t}$ need not have a zero robot-state block, since it also includes the change in physical response between historical and current initial conditions.
\end{proof}

Conditioned on $(s_t,\eta_t)$, the function $a\mapsto J_{\mathrm{pert}} f(a;s_t,\eta_t)$ is a common local response map that may be nonlinear in action. Historical triplets provide approximate input--output observations of this map at the executed commands. The error bound separates finite-transition linearization from drift in the initial conditions, including both the representation Jacobian and the physical response. A short history favors this approximation when recent transitions remain near the current operating point; temporal proximity alone does not ensure this condition. Small transitions alone also do not eliminate the condition-drift term $\|J_{\mathrm{pert}}\|L_f\|\xi_i-\xi_t\|$. Shorter histories are not always preferable, since very short windows may contain too few transitions to characterize recent execution responses.

The observed robot state does not, in general, fully determine the underlying scene configuration. To account for additional query noise, partition the local Jacobian as $J_{\mathrm{pert}}=[J_{\mathrm{pert}}^s,J_{\mathrm{pert}}^\eta]$ and replace the ideal observation model by
\[
\begin{bmatrix}
s_i\\
H_i^{\mathrm{sam}}
\end{bmatrix}
=\Phi(\xi_i)+
\begin{bmatrix}
0\\n_i
\end{bmatrix},
\]
where $n_i$ denotes query representation noise. This gives the observed relation
\begin{equation}
\begin{bmatrix}
\Delta s_i\\
\Delta H_i^{\mathrm{sam}}
\end{bmatrix}
=J_{\mathrm{pert}}^s\Delta s_i+J_{\mathrm{pert}}^\eta\Delta\eta_i+
\begin{bmatrix}
0\\\Delta n_i
\end{bmatrix}
+r^{\Phi}_{i,t}.
\label{eq:app-unobserved-response}
\end{equation}
Here, $r^{\Phi}_{i,t}$ is the Taylor remainder of the noise-free map $\Phi$, rather than the total reference-condition error $\rho_{i,t}$ in Proposition~\ref{prop:action-response}.
Object motion, changes in imaging conditions, and measurement noise cannot in general be neglected without additional assumptions. If an unmodeled disturbance $w_i$ enters the physical transition as
$\Delta\xi_i=f(a_i;s_i,\eta_i)+w_i$, the response relative to the current reference conditions also contains $J_{\mathrm{pert}}w_i$, with the representation remainder evaluated along the realized transition. The action--response triplets record commanded inputs and realized outcomes. They do not attribute every observed change solely to the executed action, nor do they identify the full Jacobian or transition function. Their role in SGMR is to supply observed execution context. Likewise, temporal differences provide local change measurements; they are neither inherently denoised nor guaranteed to correspond to fixed physical points. When abrupt contact events or visibility changes violate the single smooth-neighborhood assumption, the local expansion should instead be interpreted piecewise over regimes in which the smoothness conditions remain valid.

The two propositions serve distinct roles: the first motivates an additive parameterization for a change of the ideal generative field, and the second relates nearby historical responses to a common current-condition response map with bounded approximation error. Neither establishes that a short history identifies $J$ or that the learned encoder can recover every required correction. In Eq.~\eqref{eq:motion-residual}, the full gated residual $g_{t,\tau}\odot\Delta v_{t,\tau}$ plays the role of the learned correction. Neither it nor the nominal prediction $v_{t,\tau}$ is assumed to equal its ideal counterpart in Eq.~\eqref{eq:residual-interpretation}; their jointly trained decomposition need not be unique. Whether this history conditioning improves action adjustment is evaluated empirically. The local analysis does not guarantee closed-loop stability or robustness to arbitrarily large perturbations.

\section{Details of Compared Models}
\label{app:compared-models}

We summarize the problems and core designs of the baseline methods
included in our experiments.

\paragraph{OpenVLA.}
OpenVLA~\citep{kim2024openvla} provides an open-source generalist policy
that can be adapted to new manipulation tasks with limited computing
resources. It combines a Llama~2 backbone with fused DINOv2 and SigLIP
visual features, and learns to predict discrete action tokens from
diverse robot demonstrations.

\paragraph{OpenVLA-OFT.}
OpenVLA-OFT~\citep{kim2025fine} studies how to adapt pretrained VLAs
to new robot setups while improving control performance and inference
speed. Its optimized fine-tuning recipe combines parallel decoding,
action chunking, continuous action outputs, and an $L_1$ regression
objective.

\paragraph{CogVLA.}
CogVLA~\citep{li2026cogvla} targets the computational cost of processing
dense visual tokens in VLA models. It uses instruction-conditioned
visual aggregation and token pruning to retain task-relevant
information, together with an attention scheme that couples causal
vision-language processing and parallel action decoding.

\paragraph{$\pi_0$.}
$\pi_0$~\citep{black2024pi_0} connects pretrained vision-language
knowledge with continuous control for dexterous manipulation.
It uses a flow-matching action expert to generate action chunks
conditioned on a VLM backbone, with broad training across robot
platforms followed by task-specific fine-tuning.

\paragraph{$\pi_{0.5}$.}
$\pi_{0.5}$~\citep{intelligence2025pi_} focuses on generalization
to environments and objects outside the robot training distribution.
It combines heterogeneous robot data, semantic prediction tasks,
and web data in a co-training recipe that connects high-level task
understanding with continuous action generation.

\paragraph{VLA-Adapter.}
VLA-Adapter~\citep{wang2026vla} studies efficient adaptation from
vision-language representations to robot actions without large-scale
robotic pretraining. Its lightweight policy module uses Bridge
Attention to inject relevant vision-language features into action
generation, reducing the cost of connecting a pretrained VLM to control.

\paragraph{MemoryVLA.}
MemoryVLA~\citep{shi2026memoryvla} addresses the limited temporal
context available to policies that primarily use the current
observation. It maintains perceptual details and cognitive semantics
in a memory bank, retrieves relevant historical entries, and fuses
them with current features to condition a diffusion action expert.

\paragraph{X-VLA.}
X-VLA~\citep{zheng2026x} addresses differences in observation and
action spaces across robot embodiments. It conditions a shared
flow-matching transformer on learned embodiment-specific soft
prompts, allowing the policy to use heterogeneous robot data and
adapt to new embodiments.

\paragraph{Evo-1.}
Evo-1~\citep{lin2026evo} targets the computational cost of large VLAs
and the degradation of vision-language representations during robot
fine-tuning. With 0.77B parameters, it adds a cross-modulated
diffusion transformer to a native multimodal VLM and first aligns the
action module with the frozen VLM before fine-tuning the full model.

\paragraph{GeoPredict.}
GeoPredict~\citep{qian2026geopredict} addresses the reactive,
2D-centric behavior of VLAs in tasks that require precise 3D
reasoning. It encodes motion history, predicts multi-step 3D keypoint
trajectories of the robot arm and future workspace geometry as 3D
Gaussians, and uses these predictions only as training-time
supervision through depth rendering, so inference adds only query
tokens.

\paragraph{AVA-VLA.}
AVA-VLA~\citep{xiao2026ava} addresses the history-agnostic processing
of observations in most VLAs. It treats manipulation as a partially
observable decision process, maintains a recurrent state that
approximates the belief over the task history, and uses this state to
reweight the visual tokens of the current observation. We report its
results with one policy for all four suites.

\paragraph{AAC.}
Adaptive Action Chunking (AAC)~\citep{liang2026adaptive} addresses the
trade-off between reactivity and consistency imposed by a fixed
action-chunk length. At inference, it samples several candidate
chunks, estimates the action entropy, and chooses the number of
executed actions accordingly. We report its results with the
$\pi_{0.5}$ backbone.

\paragraph{QuantVLA.}
QuantVLA~\citep{zhang2026quantvla} addresses the memory and compute
cost of deploying VLAs. This training-free post-training quantization
framework integerizes the linear layers of the language backbone and
the diffusion-transformer action head, keeps attention projections in
floating point, and calibrates attention temperature and output
energy. We report $\pi_{0.5}$ with W4A8 quantization.

\paragraph{$\pi_0$-FAST.}
$\pi_0$-FAST~\citep{pertsch2025fast} addresses the difficulty of
learning high-frequency actions with independently discretized
action values. It uses discrete cosine transforms and byte-pair
encoding to compress action chunks into tokens, enabling
autoregressive action prediction with shorter target sequences.

\paragraph{Spatial Forcing.}
Spatial Forcing~\citep{li2026spatial} addresses limited spatial
awareness in VLAs built on vision-language backbones pretrained
primarily on 2D images. It aligns intermediate visual embeddings
with geometric features from pretrained 3D foundation models,
providing spatial supervision without requiring explicit 3D
observations as policy inputs.

\paragraph{SRPO.}
SRPO~\citep{fei2026srpo} addresses sparse success rewards that
overlook partial progress in failed rollouts. It uses successful
trajectories from the current training batch as references and
compares trajectories in a world model's latent space to assign
progress-based rewards for policy optimization.

\paragraph{MergeVLA.}
MergeVLA~\citep{fu2026mergevla} addresses interference between
task-specific updates when combining independently trained VLA
experts. It uses task-masked LoRA and action experts composed of
cross-attention-only blocks to facilitate model merging, with a
task router selecting the appropriate mask and action head.

\paragraph{VLA-JEPA.}
VLA-JEPA~\citep{sun2026vla} addresses appearance bias and reliance
on future observations when inferring latent actions from video.
It predicts future representations from the current observation
while using future frames only as supervision, then fine-tunes
an action head to connect the learned representations with control.

\paragraph{RIPT-VLA.}
RIPT-VLA~\citep{tan2025interactive} studies how demonstration-trained
VLA policies can improve through environment interaction when
expert data are scarce. It fine-tunes policies with binary success
rewards, combining dynamic rollout sampling and leave-one-out
advantage estimation for policy optimization.

\paragraph{ACoT-VLA.}
ACoT-VLA~\citep{zhong2026acot} addresses the limited action detail
conveyed by intermediate language descriptions or synthesized goal
images. It combines explicit coarse reference trajectories with
implicit action priors extracted from multimodal representations,
using both forms of action-space reasoning to condition the action head.

\section{Implementation Details}
\label{app:implementation}

\begin{figure}[!htb]
\centering
\includegraphics[width=0.9\linewidth]{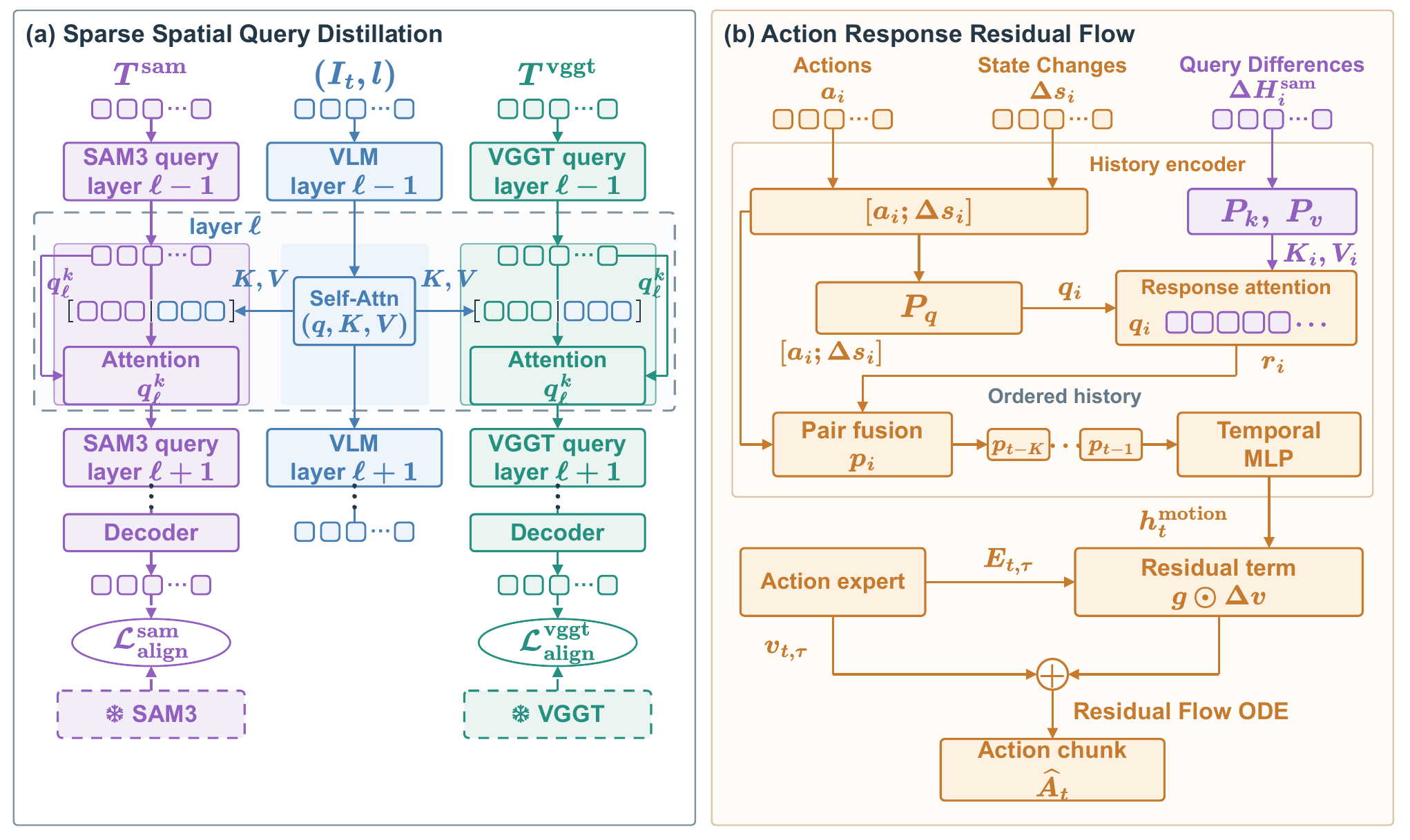}
\caption{\textbf{Spatial Supervision and Action Adjustment.}
(a) Query attention and dense teacher-feature reconstruction.
(b) History encoding and gated residual heads for action generation.
Uppercase $T$ denotes learnable initial tokens and lowercase $q$ denotes projected attention queries; time indices on
$g$ and $\Delta v$ are omitted for clarity.}
\label{fig:component-details}
\end{figure}

Table~\ref{tab:training-model-configuration} lists the training and model configurations of SGMR, and Figure~\ref{fig:component-details} illustrates the two modules.

\begin{table}[!htb]
  \centering
  \caption{\textbf{Training and model configurations of SGMR.}}
  \label{tab:training-model-configuration}
  \renewcommand{\arraystretch}{1.12}
  \small\setlength{\tabcolsep}{3pt}
  \begin{tabularx}{\linewidth}{@{}Xl@{}}
  \toprule
  \textbf{Parameter} & \textbf{Setting} \\
  \midrule
  \multicolumn{2}{@{}l}{\textit{Training configuration}} \\
  Training framework & OpenPI (JAX) \\
  Number of GPUs & 8 \\
  FSDP devices & 2 \\
  Global batch size & 128 \\
  Training steps & 30,000 \\
  Data loader workers & 8 \\
  Optimizer & AdamW \\
  Peak learning rate & $5 \times 10^{-5}$ \\
  Learning-rate schedule & Warmup cosine decay \\
  Warmup steps & 10,000 \\
  Decay steps & 1,000,000 \\
  Final learning rate & $5 \times 10^{-5}$ \\
  AdamW coefficients & $\beta_1=0.9,\ \beta_2=0.95$ \\
  Weight decay & $1 \times 10^{-10}$ \\
  Gradient clipping norm & 1.0 \\
  EMA decay & 0.999 \\
  \midrule
  \multicolumn{2}{@{}l}{\textit{Model configuration}} \\
  Action prediction horizon $H$ & 10 \\
  History transitions $K$ & 3 \\
  Teacher streams & SAM3 and VGGT \\
  Camera slots per teacher (main, wrist, padded) & 3 \\
  Tokens per camera slot per teacher & 16 \\
  Total tokens per teacher $M_k$ & 48 \\
  Query hidden dimension $d_k$ & 512 \\
  SAM3 target feature dimension $c_0$ & 1024 \\
  VGGT target feature dimension $c_1$ & 2048 \\
  SAM3 alignment loss coefficient $\lambda_{\mathrm{sam}}$ & 0.4 \\
  VGGT alignment loss coefficient $\lambda_{\mathrm{vggt}}$ & 0.1 \\
  \bottomrule
  \end{tabularx}
\end{table}

\subsection{Spatial Query Attention and Reconstruction}
\label{app:query-implementation}

\paragraph{Layerwise query attention.}
Let $L$ be the number of transformer layers and $\ell\in\{1,\ldots,L\}$
the layer index. As in Section~\ref{sec:method:query}, $k\in\{0,1\}$ indexes SAM3 and VGGT, respectively. Descriptive superscripts identify the same branches: $H_t^{\mathrm{sam}}\equiv H_t^0$ and $H_t^{\mathrm{vggt}}\equiv H_t^1$, with the same convention for $T^k$ and $\mathcal{L}_{\mathrm{align}}^k$.
The VLM context at observation step $t$ in Section~\ref{sec:method:query} is
$\mathcal{C}_t=\{\mathcal{C}_{t,\ell}\}_{\ell=1}^{L}$, where
$\mathcal{C}_{t,\ell}$ supplies the VLM keys
$K_{t,\ell}^{\mathrm{vlm}}$ and values $V_{t,\ell}^{\mathrm{vlm}}$.
Write $H_{t,0}^k=T^k$ for the initial tokens and $H_{t,L}^k=H_t^k$ for the final output. At layer $\ell$, branch $k$ projects $H_{t,\ell-1}^k$ into
attention queries $q_{t,\ell}^k$, keys $K_{t,\ell}^k$, and values
$V_{t,\ell}^k$. Lowercase $q$ denotes an attention projection;
uppercase $T^k$ denotes the learnable initial tokens. The attention output is
\begin{equation}
O_{t,\ell}^k=
\operatorname{Attn}\left(
q_{t,\ell}^k,
[K_{t,\ell}^k;K_{t,\ell}^{\mathrm{vlm}}],
[V_{t,\ell}^k;V_{t,\ell}^{\mathrm{vlm}}]
\right),
\label{eq:query-attention}
\end{equation}
where $\operatorname{Attn}$ is scaled dot-product attention and
$[\,;\,]$ concatenates tokens. The transformer block uses
$O_{t,\ell}^k$ to update the query states. The attention mask permits
each query branch to read itself and the VLM, but blocks attention
between the two query branches and from the VLM to the queries.
The action expert reads cached keys and values from all three streams
at corresponding layers.

\paragraph{Dense reconstruction and alignment loss.}
In Eqs.~\eqref{eq:spatial-query}--\eqref{eq:teacher-reconstruction},
$M_k$ and $N_k$ count all query tokens and teacher target locations,
respectively, across configured camera groups. In the LIBERO implementation, these groups include a padded camera slot in addition to the two valid camera views; the masks below select valid reconstruction targets.
The initial tokens $T^k$ and final states $H_t^k$ both have shape $M_k\times d_k$. The final states
retain $M_k$ tokens, while the teacher-specific decoder $D_{\psi_k}$ expands them
to $N_k$ predictions. For each camera, learned output slots with
two-dimensional positional embeddings attend to projected query
states from that view. Each slot predicts the feature at one target
grid location. The reconstructed and teacher grids $\widehat{F}_t^k$ and $F_t^k$ both have shape $N_k\times c_k$. Thus, $\widehat{F}_{t,j}^k$ is a decoder prediction and
$F_{t,j}^k$ is the frozen teacher feature at spatial location $j$, both with
teacher feature dimension $c_k$. We use $j$ for spatial locations and reserve $i$ for historical transitions.

Let $m_{t,j}^k\in\{0,1\}$ mark valid teacher locations, excluding
padding and invalid camera views. The alignment objective is
\begin{equation}
\mathcal{L}_{\mathrm{align}}^k
=\mathbb{E}\left[
\frac{
\sum_{j=1}^{N_k} m_{t,j}^k
\left[
1-\operatorname{cos}
\left(\widehat{F}_{t,j}^k,F_{t,j}^k\right)
\right]
}{
\max\left(1,\sum_{j=1}^{N_k}m_{t,j}^k\right)
}
\right],
\label{eq:query-reconstruction}
\end{equation}
where $\operatorname{cos}$ denotes cosine similarity. We first average
over each example's valid locations, with the denominator clamped to
at least one, then take the expectation over training examples.
The objective compares reconstructed features with teacher targets;
it does not directly match the shared VLM states to teacher features.

\subsection{Action Response Encoding and Residual Heads}
\label{app:motion-implementation}

\paragraph{History encoder.}
At replanning step $t$, the encoder uses the $K$ past transitions
$i=t-K,\ldots,t-1$, requiring $K+1$ observations including the current one. Each pairs executed action $a_i$ with the robot-state
and SAM3 query-state changes $\Delta s_i$ and $\Delta H_i^{\mathrm{sam}}$ defined in
Section~\ref{sec:method:motion}. Learned projections
$P_q$, $P_k$, and $P_v$ form
$q_i=P_q([a_i;\Delta s_i])$, $K_i=P_k(\Delta H_i^{\mathrm{sam}})$, and
$V_i=P_v(\Delta H_i^{\mathrm{sam}})$, respectively. Their attention output is
\begin{equation}
r_i=\operatorname{Attn}(q_i,K_i,V_i).
\label{eq:motion-attention}
\end{equation}
Here, $r_i$ summarizes the query-state response conditioned on the
action and robot-state change; $K_i$ denotes attention keys, distinct
from the history length $K$. We form transition descriptors
$p_i=f_{\mathrm{pair}}([r_i;a_i;\Delta s_i])$ and the execution context
$h_t^{\mathrm{motion}}=f_{\mathrm{temp}}([p_{t-K};\ldots;p_{t-1}])$.
Both $f_{\mathrm{pair}}$ and $f_{\mathrm{temp}}$ are MLPs with input
layer normalization; $[\,;\,]$ here concatenates features and, for
$f_{\mathrm{temp}}$, preserves temporal order. The context is computed once
per replanning step and reused throughout the generated action chunk.

\paragraph{Residual and gate.}
Let $E_{t,\tau}\in\mathbb{R}^{H\times d_e}$ contain the final
action-expert features at flow time $\tau$, where $H$ is the action
chunk length and $d_e$ the expert feature dimension. A learned
projection $P_{\mathrm{m}}$ maps the execution context to $d_e$
dimensions and broadcasts it across the $H$ action tokens:
\begin{equation}
\begin{aligned}
U_{t,\tau}
&=E_{t,\tau}+\mathbf{1}_H[P_{\mathrm{m}}(h_t^{\mathrm{motion}})]^\top,\\
\Delta v_{t,\tau}&=f_{\mathrm{res}}(U_{t,\tau}),\qquad
g_{t,\tau}=\sigma(f_{\mathrm{gate}}(U_{t,\tau})).
\end{aligned}
\label{eq:motion-heads}
\end{equation}
Here, $\mathbf{1}_H$ is a length-$H$ column of ones, $U_{t,\tau}$
contains the conditioned features, $f_{\mathrm{res}}$ and
$f_{\mathrm{gate}}$ are lightweight MLP heads, and $\sigma$ is the
elementwise sigmoid. The nominal velocity $v_{t,\tau}$, residual
$\Delta v_{t,\tau}$, gate $g_{t,\tau}$, and corrected velocity
$\widetilde{v}_{t,\tau}$ all have shape $H\times d_a$, matching the
action chunk $A_t$. The residual and gate modify the nominal field
as specified in Eq.~\eqref{eq:motion-residual}.
This velocity belongs to the generative flow over action chunks,
rather than denoting physical robot velocity.

\paragraph{Training.}
The VLM backbone and action expert are initialized from pretrained
$\pi_{0.5}$ weights. VGGT and SAM3 remain frozen. Each query branch
has independent parameters and a teacher-specific reconstruction
decoder. Query states receive supervision from both teacher-feature
reconstruction and action prediction, whereas the reconstruction
decoders are optimized only by their respective reconstruction
objectives. The gated residual is added to the nominal velocity as in
Eq.~\eqref{eq:motion-residual}, without an additional scaling coefficient,
in both training and inference.

\paragraph{Inference.}
The noisy action chunk
$z_\tau$ is embedded into the action stream, while flow time $\tau$
conditions adaptive normalization in the action expert.
Teacher encoders and reconstruction decoders are discarded.
The VLM context, query representations, and action--response
execution context are computed once per replanning step and reused
throughout action-chunk generation. The residual and gate are
recomputed at each flow step from the current action-expert states,
as described in Section~\ref{sec:method:motion}.

\section{Additional Temporal Ablations}
\label{app:temporal-ablation}

\paragraph{Temporal Context and Camera Views.}
Table~\ref{tab:ablation-motion} varies the number $K$ of past
transitions and the camera views used for action--response
conditioning. With both main and wrist views, increasing $K$ from
one to three raises success from 86.1\% to 87.6\%; $K=5$ gives
the same 87.6\%. With $K=3$, main-view-only and wrist-only history
reach 86.4\% and 86.5\%, below the 87.6\% obtained with both views.
These results favor a short recent history in this setting.

\begin{table}[!htb]
\centering
\caption{\textbf{Action Response Residual Flow ablations on LIBERO-Plus.}
We vary the number $K$ of past transitions (the residual flow
receives the $K$ executed actions and $K+1$ observations, including
the current one) and the camera views used for motion conditioning.
Main and Wrist denote the main and wrist-mounted cameras,
respectively.
Success rates (SR) are reported as percentages (\%).}
\label{tab:ablation-motion}
\small
\setlength{\fboxsep}{1pt}
\setlength{\tabcolsep}{14pt}
\renewcommand{\arraystretch}{1.12}
\begin{tabular}{@{}ccc@{}}
\toprule
$\boldsymbol{K}$
 & \textbf{Camera}
 & \textbf{SR (\%)} \\
\midrule
1 & Main + Wrist & 86.1 \\
3 & Main + Wrist & \colorbox{lightred}{87.6} \\
5 & Main + Wrist & \colorbox{lightred}{87.6} \\
\midrule
3 & Main & 86.4 \\
3 & Wrist & 86.5 \\
\bottomrule
\end{tabular}
\end{table}

\FloatBarrier
\subsection{Temporal Conditioning and Reconstruction Supervision}
\label{app:history-supervision}

\begin{table*}[!t]
\centering
\small\setlength{\fboxsep}{1pt}
\caption{\textbf{LIBERO-Plus success rates by perturbation (\%).} Cam., Rob., Lang., and BG.\ denote camera viewpoint, robot initial state, language instruction, and background perturbations, as in Table~\ref{tab:libero-plus}. Overall is the reported pooled success rate. The first group reports the retrained $\pi_{0.5}$ baseline and the dual-query models with and without the reconstruction loss. The second group compares temporal conditioning variants built on the dual-query model with reconstruction.}
\label{tab:supp-history-supervision}
\begin{tabular}{@{}lrrrrrrrr@{}}
\toprule
Configuration & Cam. & Rob. & Lang. & Light & BG. & Noise & Layout & Overall \\
\midrule
\multicolumn{9}{@{}l}{\emph{Spatial query supervision}} \\
$\pi_{0.5}$ (retrained) & 69.5 & 75.7 & 82.1 & 96.3 & \colorbox{lightred}{94.8} & 79.1 & 85.9 & 82.2 \\
Dual queries, no reconstruction & 71.4 & 73.5 & 82.4 & \colorbox{lightred}{97.5} & 94.5 & 83.8 & \colorbox{lightred}{87.2} & 83.2 \\
Dual queries, with reconstruction & \colorbox{lightred}{74.7} & \colorbox{lightred}{79.9} & \colorbox{lightred}{84.8} & 95.6 & 92.6 & \colorbox{lightred}{90.7} & \colorbox{lightred}{87.2} & \colorbox{lightred}{85.8} \\
\midrule
\multicolumn{9}{@{}l}{\emph{Temporal conditioning}} \\
Standard history conditioning & 68.4 & 84.8 & 81.7 & 95.4 & \colorbox{lightred}{95.1} & 91.1 & 87.5 & 85.4 \\
Residual flow, action + proprio. & 72.0 & 86.0 & 82.8 & 94.0 & 92.1 & 89.6 & 87.3 & 85.6 \\
Residual flow, visual + proprio. & 72.0 & 78.1 & 83.7 & \colorbox{lightred}{96.8} & 93.9 & 90.4 & 87.3 & 85.2 \\
Residual flow, action + visual & 72.5 & 86.3 & 85.3 & 96.5 & 94.1 & 92.3 & \colorbox{lightred}{88.3} & 87.2 \\
Residual flow, shuffled action response & 71.7 & 84.8 & 85.2 & 96.6 & 92.2 & 89.5 & 87.0 & 86.0 \\
\textbf{SGMR} & \colorbox{lightred}{75.0} & \colorbox{lightred}{86.6} & \colorbox{lightred}{85.9} & 94.8 & 94.1 & \colorbox{lightred}{93.0} & 87.9 & \colorbox{lightred}{87.6} \\
\bottomrule
\end{tabular}

\end{table*}

Table~\ref{tab:supp-history-supervision} extends Table~\ref{tab:history-supervision} to all seven LIBERO-Plus dimensions. The three partial-input variants keep the residual branch and drop one element of the triplet: action + proprio.\ drops the visual responses, visual + proprio.\ drops the executed actions, and action + visual drops the robot-state changes. Between the first two, the difference is largest under robot initial-state changes (86.0\% vs.\ 78.1\%), where the executed actions matter most; under lighting shifts, the visual + proprio.\ variant scores higher (96.8\% vs.\ 94.0\%). Dropping the robot-state changes costs the least, 0.4 points overall (87.2\% vs.\ 87.6\%). Shuffling the pairing of actions and responses lowers success under camera (71.7\% vs.\ 75.0\%) and robot initial-state (84.8\% vs.\ 86.6\%) perturbations and overall (86.0\% vs.\ 87.6\%), consistent with Section~\ref{sec:experiments:ablation}. Reconstruction supervision mainly improves robot initial-state (73.5\% to 79.9\%) and noise (83.8\% to 90.7\%) perturbations.

\section{Action Adjustment under Perturbations}
\label{app:action-adjustment}

We examine Action Response Residual Flow through matched LIBERO-Plus
rollouts. Base denotes the retrained $\pi_{0.5}$ policy (Baseline in Figure~\ref{fig:ablation-module}), Query adds both query branches (VQ + SQ), and Query + Residual Flow further adds the
action response encoder and residual flow, giving the full SGMR model. For this component comparison, the three policies share the task, instruction,
perturbation, and simulator initial state. We visualize intermediate
action changes and each rollout's actual endpoint. Numbers denote control
steps, blue outlines highlight adjustments, and the last column shows
each actual endpoint. Checks and crosses follow the logged full-task
success criteria.

\begin{figure}[!htbp]
\centering
\includegraphics[width=\linewidth]{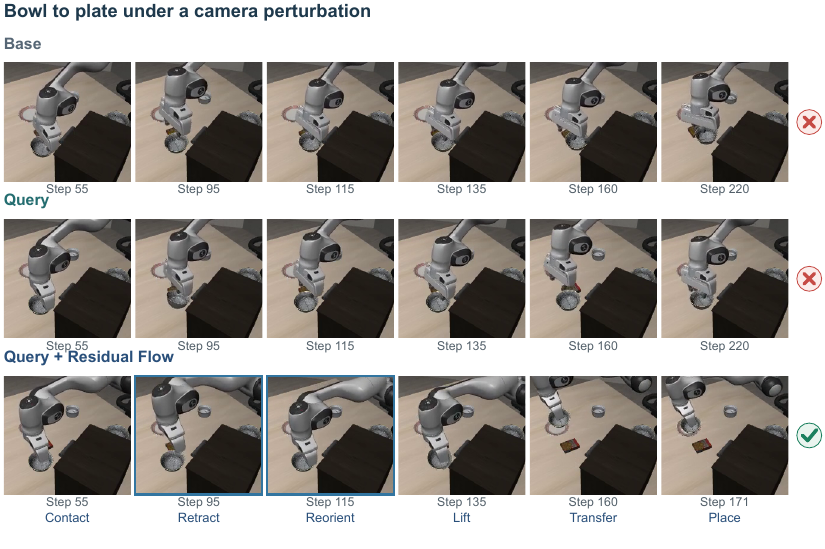}
\caption{\textbf{Camera perturbation: post-contact adjustment.}
Blue outlines highlight retraction and reorientation after initial contact.}
\label{fig:motion-adjustment}
\end{figure}

\paragraph{Correction after contact.}
After initial contact without lifting the bowl, Query + Residual Flow
retracts, reorients, and completes the transfer at step 171
(Figure~\ref{fig:motion-adjustment}). Base and Query also adjust their
poses but do not finish by step 220.

\begin{figure}[!htbp]
\centering
\includegraphics[width=\linewidth]{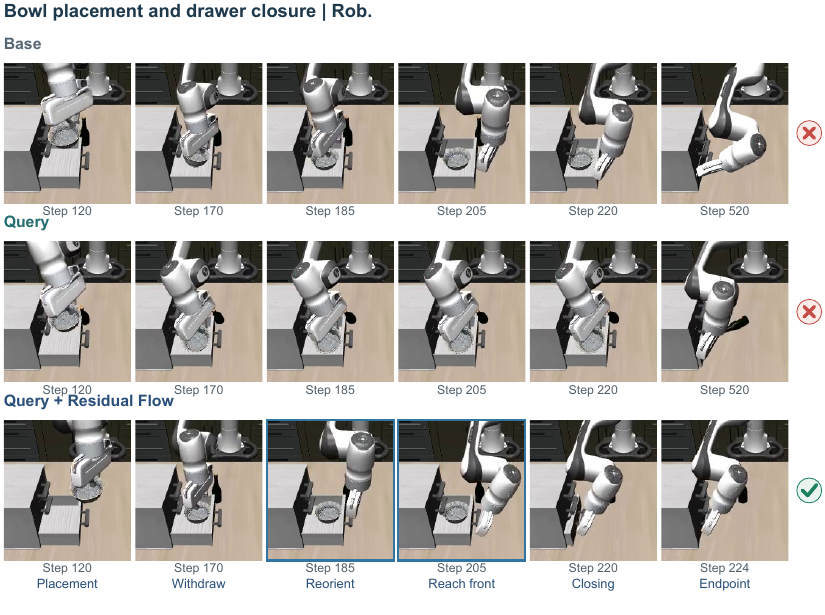}
\caption{\textbf{Robot-initial-state perturbation: placement-to-closing transition.}
Blue outlines mark gripper reorientation and the approach to the drawer front.}
\label{fig:residual-flow-robot}
\end{figure}

\paragraph{Adjustment between task stages.}
Figure~\ref{fig:residual-flow-robot} shows the task of placing a bowl
in the bottom drawer and then closing it under a robot-initial-state
perturbation. After placing the bowl, the residual-flow variant
withdraws the gripper, reorients toward the drawer front, and completes
the task at step 224. The query-only variant remains close to the bowl
during the corresponding trajectory segment, while neither Base nor
Query satisfies the full-task success criterion within 520 steps.
This example illustrates adjustment between successive task stages.

\begin{figure}[!htbp]
\centering
\includegraphics[width=\linewidth]{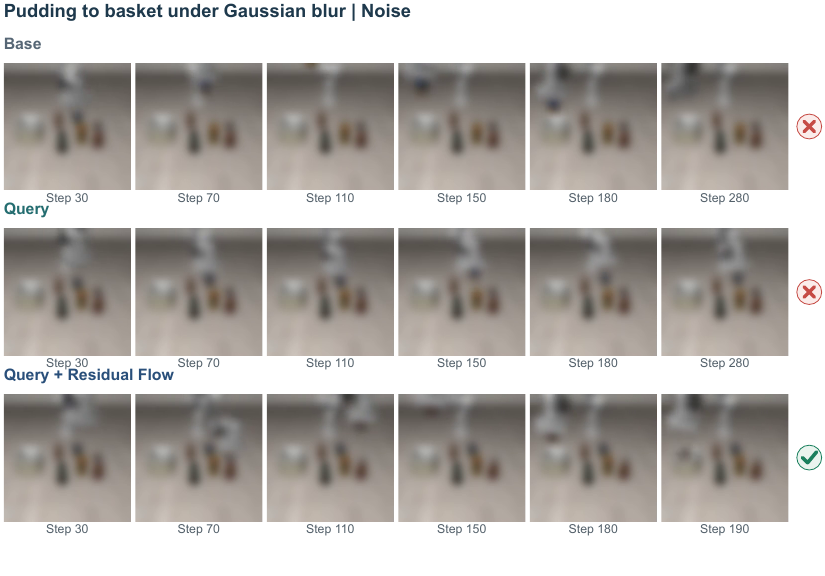}
\caption{\textbf{Noise perturbation: execution under Gaussian blur.}
Matched rollouts for placing the chocolate pudding in the basket under
LIBERO-Plus \texttt{noise\_19} ($\sigma=9$ pixels).
The original blurred observations are shown without additional processing.}
\label{fig:residual-flow-noise}
\end{figure}

\paragraph{Execution with degraded visual observations.}
In Figure~\ref{fig:residual-flow-noise}, Query + Residual Flow completes
the transfer at step 190, whereas Base and Query do not satisfy the
task-success criterion within 280 steps. Severe blur obscures fine
contact events, so this case supports successful execution with
degraded visual input without identifying a specific recovery
mechanism. Together, the three selected examples complement the
aggregate component ablations in Figure~\ref{fig:ablation-module};
they illustrate execution behavior rather than estimate overall
success rates or establish causal attribution.

\section{Real-World Evaluation Details}
\label{app:real-world-recordings}

\paragraph{Robot platform.}
We use one Synria Robotics Alicia-D follower arm with six revolute joints
and a 50\,mm-stroke parallel-jaw gripper. The manufacturer specifies
a 650\,mm working radius and $\pm2$\,mm repeatability.\footnote{\href{https://docs.sparklingrobo.com/en/docs/alicia-d-series}{Synria Robotics, \emph{Alicia-D Series}, technical specifications.}}
An external RGB camera overlooks the tabletop and a second is wrist-mounted.
A separate leader arm sets the follower's initial pose before policy execution.

\begin{samepage}
\paragraph{Demonstrations.}
We used the instructions in Table~\ref{tab:alicia-task-details} for all
150 nominal demonstrations (50 per task) and all four evaluation conditions.\par
\end{samepage}

\begin{table}[htbp]
  \centering
  \small
  \setlength{\tabcolsep}{5pt}
  \renewcommand{\arraystretch}{1.15}
  \caption{Demonstrations and policy instructions for the three Alicia-D tasks.}
  \label{tab:alicia-task-details}
  \begin{tabularx}{\linewidth}{@{}p{0.25\linewidth}cX@{}}
    \toprule
    Task & Demos & Language instruction \\
    \midrule
    Block stacking & 50 & ``Stack the red block on the green block.'' \\
    Block-to-plate placement & 50 & ``Place the green block into the green plate.'' \\
    Earphones-to-box placement & 50 & ``Put the earphones into the yellow box.'' \\
    \bottomrule
  \end{tabularx}
\end{table}

\paragraph{Observations and control.}
We resized main- and wrist-camera RGB frames from $1280\times720$ to
$224\times126$ using Lanczos resampling and padded 49 pixels above and below.
The Alicia-D SDK reads joint angles and gripper opening over USB serial
and computes the tool pose in the arm base frame by forward kinematics.
Our hardware check used SDK v6.1.0 and firmware v6.1.3. Gripper feedback
ranges from 0 (closed) to 1000 (open).
Each policy state and action contains three Cartesian translation coordinates,
three rotation-vector components, and one gripper coordinate. The controller converts every
relative target in an action chunk to an absolute pose using the same
end-effector pose measured at observation time. Commands within a segment
are dispatched at a nominal 15\,Hz. During synchronous replanning, the
controller holds the last target until the next prediction arrives.
The cameras captured approximately 7.2--7.3 frames per second in the
selected recordings.

\paragraph{Perturbations.}
We varied the main camera's mounting pitch over $90^\circ$--$120^\circ$
and yaw over $-60^\circ$--$60^\circ$. Layout trials varied object and
receptacle positions. For visual noise, we used motion blur, Gaussian blur, zoom blur, fog,
and glass blur from LIBERO-Plus~\citep{fei2025libero}. For each
corruption we sampled three distinct severity levels out of ten using
seed 20261111 and applied the same 15 configurations to both policies.
We corrupted only the unpadded main-camera image. Within a configuration,
the motion-blur direction, fog field, or glass-blur mapping was fixed across frames.

\paragraph{Success criterion and trial counts.}
We manually scored a trial as successful when the object reached the final
placement specified by the task instruction. Each model was evaluated in
90 trials per condition, pooled across the three tasks.
Table~\ref{tab:alicia-success-counts} gives the counts underlying the
rates in Figure~\ref{fig:real-world-overview}.

\begin{table}[htbp]
  \centering
  \small
  \renewcommand{\arraystretch}{1.12}
  \caption{\textbf{Real-world successful-trial counts and success rates.}
  Each cell gives successes/trials (percentage) for one model and
  condition, pooled across the three tasks.}
  \label{tab:alicia-success-counts}
  \begin{tabular}{@{}lcc@{}}
    \toprule
    Condition & SGMR & $\pi_{0.5}$ \\
    \midrule
    Nominal & 88/90 (97.8\%) & 86/90 (95.6\%) \\
    Visual noise & 82/90 (91.1\%) & 74/90 (82.2\%) \\
    Camera viewpoint & 74/90 (82.2\%) & 66/90 (73.3\%) \\
    Object layout & 84/90 (93.3\%) & 80/90 (88.9\%) \\
    \bottomrule
  \end{tabular}
\end{table}

\FloatBarrier

\FloatBarrier
\section{Language-extended Benchmark and VLM Representation Analysis}
\label{app:language-benchmark}

This section details the benchmark and the representation analysis of Section~\ref{sec:experiments:prior}. We report results for the six policies in Table~\ref{tab:language-prior}, all trained on LIBERO for 30k steps with a global batch size of 128 and evaluated without additional fine-tuning. VQ and SQ use the VGGT and SAM3 query branches, respectively. SGMR combines both branches with the Action Response Residual Flow. VGGT align and SAM3 align denote direct alignment of the VLM layer-12 image tokens to the corresponding teacher features.

\paragraph{Construction.}
The benchmark uses the 40 LIBERO tasks for which LIBERO-Plus provides language variants. Every episode runs the original scene and initial state of a task, and only the instruction changes. Ten LIBERO-Goal tasks share one scene and two LIBERO-10 tasks share another, so the instruction alone determines which task is requested; these 12 tasks form Split~A. The remaining 28 tasks, whose scene identifies the task, form Split~B. At every control step we evaluate the goal predicates of all tasks that can be checked in the scene, e.g., all ten LIBERO-Goal goals, which records the task that was actually completed. An episode ends when the scene's own task or the instructed task is completed, or at the LIBERO step limit.

\paragraph{Instruction types.}
Table~\ref{tab:language-types} lists the instruction types. The lexical, commonsense, syntactic, distractor and multilingual rewrites were written for each task so that they request the same task and refer to every object unambiguously within its scene, and were checked against the object inventory of each scene before evaluation. LP contains all 1{,}537 LIBERO-Plus rewrites of the 40 tasks. Controls replace the instruction with an empty string, a fixed pseudo-word string, or the original instruction of another task.

\begin{table*}[!t]
\centering
\caption{\textbf{Instruction types.} Counts are the number of distinct instructions per task in Split~A and Split~B; examples are for ``put the cream cheese in the bowl'' (LIBERO-Goal), except the swap controls, which are for ``pick up the milk and place it in the basket'' (LIBERO-Object).}
\label{tab:language-types}
\small
\setlength{\tabcolsep}{4pt}
\begin{tabular}{@{}llccp{0.42\linewidth}@{}}
\toprule
Type & Content & A & B & Example \\
\midrule
P0 & Original (training) instruction & 1 & 1 & \emph{``put the cream cheese in the bowl''} \\
P1 & Case, punctuation, 1--2 typos & 4 & 2 & \emph{``put the craem cheese in the bwl''} \\
P2 & Synonyms & 5 & 2 & \emph{``place the cream cheese in the bowl''} \\
P3 & Commonsense descriptions of objects & 5 & 2 & \emph{``put the soft cheese spread in the black soup dish''} \\
P4 & Goal-state or reordered phrasing & 4 & 1 & \emph{``the cream cheese should end up in the bowl''} \\
P5 & Irrelevant context or negated mentions & 4 & 1 & \emph{``I'm making breakfast, so put the cream cheese in the bowl''} \\
P6 & Chinese, Spanish, French & 3 & 3 & \emph{``pon el queso crema en el cuenco''} \\
LP & All LIBERO-Plus rewrites of the task & 19--50 & 23--49 & \emph{``could you put the cream cheese into the bowl''} \\
C & Empty instruction & 1 & 1 & \emph{(empty)} \\
C & Fixed pseudo-word string & 1 & 1 & \emph{``serot lepig pumuf tunoz fusop fegit din det senedep''} \\
C & Instruction of another task feasible in the scene & 0 & 0--2 & \emph{``pick up the cream cheese and place it in the basket''} \\
C & Instruction of a task whose target is absent & 0 & 1 & \emph{``pick up the ketchup and place it in the basket''} \\
\bottomrule
\end{tabular}

\end{table*}

\begin{table*}[!t]
\centering
\caption{\textbf{Success rate (\%) by instruction type and split.}
  Sem.\ averages P2, P3 and P6. All averages P1--P6. Blank-instruction SR measures completion of the scene's original task.
  Shading marks the highest displayed SR in each row within each split, including ties.
  On the Sem.\ and LP rows, $\uparrow$ and $\downarrow$ indicate a smaller or larger rewrite penalty than for $\pi_{0.5}$ when the paired 95\% interval excludes zero.}
\label{tab:language-dsr}
\small
\setlength{\tabcolsep}{2pt}
\setlength{\fboxsep}{1pt}
\renewcommand{\arraystretch}{1.08}
\begin{tabular*}{\linewidth}{@{\extracolsep{\fill}}lrrrrrrrrrrrr@{}}
\toprule
 & \multicolumn{6}{c}{Split A (instruction decides the task)} & \multicolumn{6}{c}{Split B (scene identifies the task)} \\
\cmidrule(lr){2-7}\cmidrule(l){8-13}
Instruction & $\pi_{0.5}$ & \shortstack{VGGT\\align} & \shortstack{SAM3\\align} & VQ & SQ & SGMR & $\pi_{0.5}$ & \shortstack{VGGT\\align} & \shortstack{SAM3\\align} & VQ & SQ & SGMR \\
\midrule
Original instruction & 98.3 & 95.8 & 97.5 & 95.8 & \colorbox{lightred}{100.0} & 99.2 & 95.2 & \colorbox{lightred}{97.6} & \colorbox{lightred}{97.6} & \colorbox{lightred}{97.6} & \colorbox{lightred}{97.6} & \colorbox{lightred}{97.6} \\
\cmidrule(lr){2-7}\cmidrule(l){8-13}
P1 surface & 65.6 & 65.6 & 65.0 & 69.0 & 71.7 & \colorbox{lightred}{73.5} & 96.4 & 94.6 & 95.8 & 95.8 & 97.0 & \colorbox{lightred}{98.8} \\
P2 lexical & 94.3 & 95.8 & 95.3 & 97.8 & \colorbox{lightred}{99.2} & 98.2 & \colorbox{lightred}{98.2} & 94.0 & 92.9 & 97.0 & \colorbox{lightred}{98.2} & 97.6 \\
P3 commonsense & 58.2 & 53.0 & 50.3 & 60.0 & 62.5 & \colorbox{lightred}{63.8} & 83.9 & 76.8 & 69.6 & 81.5 & 87.5 & \colorbox{lightred}{88.1} \\
P4 syntax & 90.0 & 91.2 & 87.5 & 94.6 & 94.6 & \colorbox{lightred}{94.8} & 96.4 & 96.4 & 95.2 & 98.8 & \colorbox{lightred}{100.0} & 96.4 \\
P5 distractor & 81.7 & 83.1 & \colorbox{lightred}{84.2} & 80.6 & 79.6 & 81.2 & 92.9 & 88.1 & 86.9 & 92.9 & \colorbox{lightred}{96.4} & 92.9 \\
P6 multilingual & \colorbox{lightred}{22.8} & 15.3 & 14.2 & 21.4 & 18.9 & 18.9 & 73.0 & 63.1 & 48.4 & 71.0 & 73.8 & \colorbox{lightred}{75.0} \\
LP LIBERO-Plus & 65.1 & 69.0 & 62.3 & \colorbox{lightred}{76.8}\textsuperscript{$\uparrow$} & 75.2 & 76.4\textsuperscript{$\uparrow$} & 89.1 & 84.1\textsuperscript{$\downarrow$} & 77.6\textsuperscript{$\downarrow$} & 89.1 & 91.1 & \colorbox{lightred}{92.4} \\
\cmidrule(lr){2-7}\cmidrule(l){8-13}
Sem. (P2, P3, P6) & 58.4 & 54.7 & 53.3 & 59.7 & 60.2 & \colorbox{lightred}{60.3} & 85.1 & 78.0\textsuperscript{$\downarrow$} & 70.3\textsuperscript{$\downarrow$} & 83.2 & 86.5 & \colorbox{lightred}{86.9} \\
All (P1--P6) & 68.8 & 67.4 & 66.1 & 70.6 & 71.1 & \colorbox{lightred}{71.7} & 90.1 & 85.5 & 81.5 & 89.5 & \colorbox{lightred}{92.2} & 91.5 \\
\cmidrule(lr){2-7}\cmidrule(l){8-13}
Blank instruction & 16.7 & 14.2 & 14.2 & \colorbox{lightred}{19.2} & 15.8 & 14.2 & \colorbox{lightred}{69.0} & 50.0 & 44.0 & 67.9 & 65.5 & \colorbox{lightred}{69.0} \\
\bottomrule
\end{tabular*}

\end{table*}

\paragraph{Protocol and statistics.}
Split~A runs every instruction from initial states 0--9 and Split~B from initial states 0--2; LP runs from initial state 0, as in LIBERO-Plus. This yields 6{,}238 episodes per model. For SGMR, each inference call also receives the frames of both cameras, the robot states and the executed actions of the last $K=3$ transitions, assembled as in our LIBERO-Plus evaluation. The flow-matching noise of each inference call is a deterministic function of the scene, the initial state and the replanning step, so all instructions and models receive identical noise and comparisons are paired. We average task success rates with equal weight across tasks. For paired tests, we subtract the success rate under rewrites from the original-instruction success rate of the same task to obtain its rewrite penalty. We use 95\% percentile intervals from 2{,}000 paired bootstrap resamples of tasks.

\begin{table*}[!t]
\centering
\caption{\textbf{Layer-wise within-frame CKA.}
  $n_+/n_-$ counts layer inputs with higher/lower CKA to the pretrained VLM than for the fine-tuned $\pi_{0.5}$ baseline, using paired 95\% task-bootstrap intervals.
  The counts cover 18 image inputs and 17 language inputs, excluding the shared language embedding.
  An interval that includes zero contributes to neither count. Shading marks the highest displayed language CKA at each reported layer, including ties.}
\label{tab:language-drift-pairs}
\small
\setlength{\tabcolsep}{3.5pt}
\setlength{\fboxsep}{1pt}
\renewcommand{\arraystretch}{1.08}
\begin{tabular*}{\linewidth}{@{\extracolsep{\fill}}l rrr rrr@{}}
\toprule
 & \multicolumn{3}{c}{Layer inputs ($n_+/n_-$)} & \multicolumn{3}{c}{Within-frame language CKA} \\
\cmidrule(lr){2-4}\cmidrule(l){5-7}
Model & Main camera & Wrist camera & Language & L5 & L12 & L16 \\
\midrule
$\pi_{0.5}$ & -- & -- & -- & 0.70 & 0.67 & 0.29 \\
VGGT align & 5/13 & 3/15 & 2/15 & 0.60 & 0.64 & 0.20 \\
SAM3 align & 0/18 & 0/17 & 2/12 & 0.68 & 0.56 & 0.24 \\
VQ & 18/0 & 15/0 & 16/0 & \colorbox{lightred}{0.72} & 0.70 & 0.31 \\
SQ & 10/4 & 7/10 & 14/1 & 0.71 & 0.70 & 0.24 \\
\midrule
SGMR & 12/1 & 10/2 & 17/0 & \colorbox{lightred}{0.72} & \colorbox{lightred}{0.71} & \colorbox{lightred}{0.37} \\
\bottomrule
\end{tabular*}

\end{table*}

\begin{table*}[!t]
\centering
\caption{\textbf{Mean within-frame CKA and relative weight change at 30k steps.}
  CKA compares each policy with the pretrained $\pi_{0.5}$ and averages over the inputs to the 18 VLM layers.
  Means follow Table~\ref{tab:language-prior} at its two-decimal precision. The SGMR wrist mean has no matched bootstrap interval.
  Weight change is the relative Frobenius norm for the attention and MLP weights. Shading marks the highest CKA or smallest weight change per column, including ties.}
\label{tab:language-drift-models}
\small
\setlength{\tabcolsep}{4pt}
\setlength{\fboxsep}{1pt}
\renewcommand{\arraystretch}{1.08}
\begin{tabular*}{\linewidth}{@{\extracolsep{\fill}}l rrr rr@{}}
\toprule
 & \multicolumn{3}{c}{Mean within-frame CKA $\uparrow$} & \multicolumn{2}{c}{Relative weight change $\downarrow$} \\
\cmidrule(lr){2-4}\cmidrule(l){5-6}
Model & Main camera & Wrist camera & Language & Gemma & SigLIP \\
\midrule
$\pi_{0.5}$ & 0.43 & 0.40 & 0.67 & \colorbox{lightred}{0.175} & 0.133 \\
VGGT align & 0.39 & 0.34 & 0.63 & 0.180 & 0.137 \\
SAM3 align & 0.27 & 0.23 & 0.64 & 0.189 & 0.146 \\
VQ & \colorbox{lightred}{0.51} & \colorbox{lightred}{0.43} & 0.70 & 0.181 & 0.136 \\
SQ & 0.45 & 0.37 & 0.69 & 0.182 & \colorbox{lightred}{0.132} \\
\midrule
SGMR & 0.45 & \colorbox{lightred}{0.43} & \colorbox{lightred}{0.72} & 0.178 & \colorbox{lightred}{0.132} \\
\bottomrule
\end{tabular*}

\end{table*}

\paragraph{Behavioral results.}
On Split~A the original instruction yields 95.8--100\% SR across the six models.
With an empty instruction they complete the scene's original task in 14.2--19.2\% of episodes
(Table~\ref{tab:language-dsr}). On Split~B, SGMR reaches 86.9\% SR on semantic rewrites
and 92.4\% on LP rewrites. Relative to $\pi_{0.5}$,
both direct alignments incur larger penalties on these two sets of rewrites.
For VQ, SQ and SGMR, the paired 95\% intervals for the penalty differences from $\pi_{0.5}$ include zero on both rewrite sets.
SGMR incurs smaller penalties than either direct alignment on both sets in Split~B.
For semantic rewrites, SGMR's penalty is 8.9 points smaller than that of VGGT alignment
(95\% CI 2.2--16.4) and 16.6 points smaller than that of SAM3 alignment (10.1--23.8).

In Split~A, VQ and SGMR reach 76.8\% and 76.4\% SR on LP rewrites, with penalties
14.3 and 10.5 points smaller than for $\pi_{0.5}$. Commonsense and multilingual rewrites are the
hardest P1--P6 types for all models. Each configuration was trained once. The intervals
quantify variation across tasks.

\paragraph{VLM features.}
All representation analyses compare the VLM of each model with that of the pretrained $\pi_{0.5}$ base model on identical inputs. Every checkpoint is loaded into the same $\pi_{0.5}$ architecture and only the prefix of images and instruction is passed through the VLM. Because prefix tokens never attend to query tokens, the query stream does not alter these features in the forward pass (maximum absolute difference 0.0 for all query checkpoints); it affects the VLM only through training. The residual flow of SGMR encodes past frames in a separate forward pass and does not alter the prefix of the current frame. We record the hidden states at the input embedding (emb) and after each of the 18 layers (L0--L17), and summaries average over emb--L16, the inputs of the 18 layers from which the keys and values attended by the action expert are computed.
\emph{Image tokens:} we sample 360 frames from the LIBERO training demonstrations (three demonstrations per task, at 15\%, 50\% and 85\% of their length), each with its training instruction, and compute linear CKA between the 256 patch tokens of one camera of a model and of the base model within each frame, with tokens centered within the frame.
\emph{Language tokens:} we use the first observation of every scene at initial states 0--2 (120 frames), paired with all instructions of its task and with the original instructions of the other tasks in the same suite (7{,}770 image--instruction pairs). Language-token states, excluding BOS, are mean-pooled, and linear CKA to the base model is computed over all instructions of a frame.
\emph{Weight change.} For the attention and MLP weights of each backbone, we compute the relative Frobenius norm $\lVert W-W_{\mathrm{base}}\rVert/\lVert W_{\mathrm{base}}\rVert$.
Per-frame values are averaged within tasks. Unless stated otherwise, intervals use 95\% task bootstraps over the 40 tasks. For within-frame CKA, the available interval endpoints lie within 0.014 of the unrounded mean.

\paragraph{Representation drift.}
Table~\ref{tab:language-drift-models} reports mean within-frame CKA for each token type.
VQ has the highest main-camera mean (0.51), followed by SQ and SGMR (both 0.45 at the displayed precision).
VQ and SGMR share the highest wrist-camera mean (0.43), compared with 0.40 for the fine-tuned baseline and 0.37 for SQ.
SGMR has the highest mean language CKA (0.72).
For all three token types, VQ and SQ have higher mean CKA than their respective direct alignments, and SGMR exceeds both direct alignments.
For main-camera and language tokens, these paired 95\% intervals favor the query policy at 15 or more layer inputs and favor direct alignment at no more than two.
Both direct alignments have lower mean CKA than the fine-tuned $\pi_{0.5}$ baseline for all three token types.
All three query policies have higher mean CKA than this baseline for main-camera and language tokens.
Table~\ref{tab:language-drift-pairs} gives the layer counts relative to the fine-tuned baseline and language CKA at L5, L12 and L16. Language-token CKA is lower at L16 than at L5 or L12 for all six policies. VQ and SGMR share the highest displayed CKA at L5 (0.72). SGMR has the highest CKA at L12 (0.71) and L16 (0.37).

Relative weight changes span 0.175--0.189 for Gemma and 0.132--0.146 for SigLIP across the six policies (Table~\ref{tab:language-drift-models}).

\section{Inference Efficiency and Model Size}
\label{app:inference-cost}

\begin{figure}[!htb]
\centering
\includegraphics[width=\linewidth]{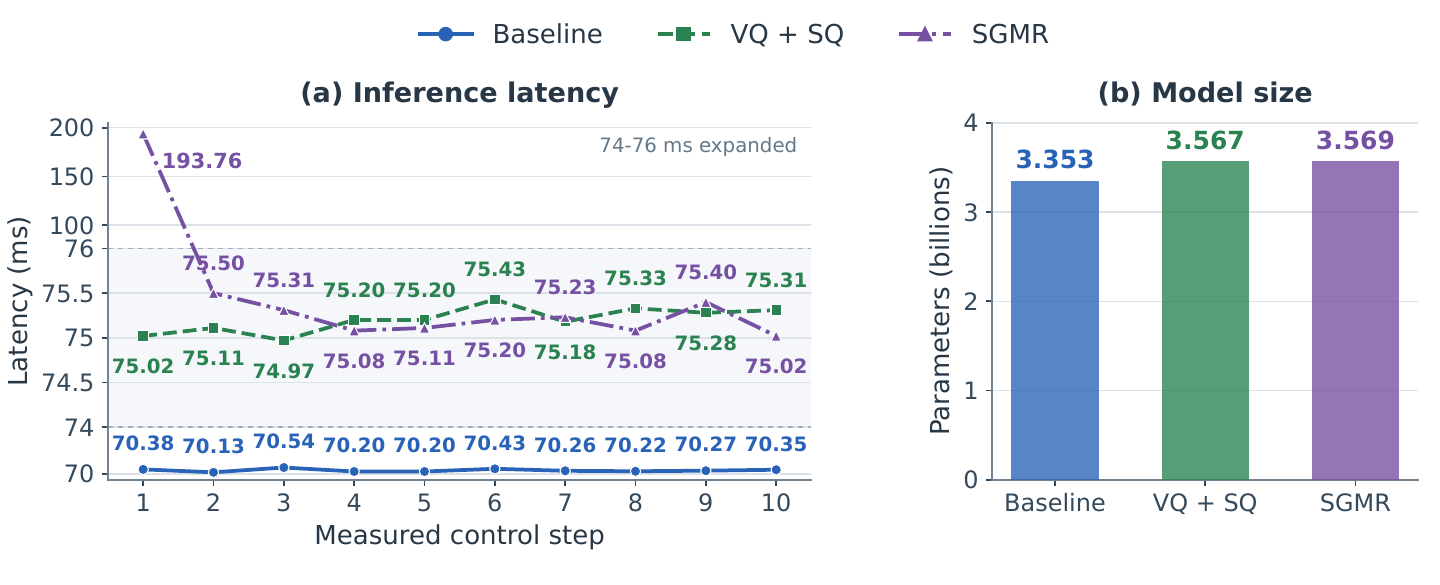}
\caption{\textbf{Inference cost on an NVIDIA H20 GPU.}
(a) Inference latency at ten control steps. The first SGMR call includes
history-cache initialization. (b) Reported parameter counts. The latency axis uses a piecewise-linear
scale that expands 74--76\,ms. The parameter axis is linear.}
\label{fig:inference-cost}
\end{figure}

\paragraph{Evaluation setting.}
We compare the $\pi_{0.5}$ baseline, the VQ + SQ configuration, and
the full SGMR model in a single-task evaluation on an NVIDIA H20 GPU.
VQ + SQ includes both the VGGT (VQ) and SAM3 (SQ) query transformers without the Action Response
Residual Flow. Figure~\ref{fig:inference-cost} reports inference latency
at ten measured control steps and the parameter count for each policy.
The first SGMR call in this trace takes 193.76\,ms and includes initialization
of the temporal-difference history cache. While the controller executes the
current action chunk, SGMR runs the next inference call asynchronously
and saves the updated history in the cache for the next replanning step.

\paragraph{Inference latency.}
Over steps 2--10, mean inference latencies are 70.29, 75.22, and
75.21\,ms per call for the baseline, VQ + SQ, and SGMR, respectively.
SGMR takes approximately 4.93\,ms (7.0\%) longer per call than the
baseline. For SGMR, the time spent waiting for the next chunk depends
on when inference finishes relative to execution of the current chunk.

\paragraph{Parameter overhead.}
The reported model sizes are 3.353B, 3.567B, and 3.569B parameters
for the baseline, VQ + SQ, and SGMR, respectively.
SGMR therefore increases the reported parameter count by approximately
6.4\% over the baseline, while adding the Action Response Residual Flow
to VQ + SQ accounts for approximately 2M parameters at the reported
precision. The query branches account for most of the parameter increase,
and the full model has a similar measured latency to VQ + SQ after the
first step in this evaluation. These are descriptive measurements from
one task, rather than estimates across repeated runs or benchmark suites.

\FloatBarrier

\section{Limitations}
\label{app:limitations}

The language and within-frame CKA analyses use one training run per policy,
with intervals estimated over tasks. At the pretrained VLM's best retrieval
layer (L8), its instruction-retrieval accuracy exceeded that of the six
fine-tuned policies on both splits. Comparisons among those policies were
inconclusive. Our real-world results pool three tabletop tasks on one
Alicia-D arm, leaving transfer to other embodiments for future evaluation.

\FloatBarrier
\end{document}